\documentclass[preprint,10pt]{elsarticle}

\usepackage[T1]{fontenc}
\usepackage[utf8]{inputenc}
\usepackage{lmodern}
\usepackage{microtype}

\usepackage{amsmath,amssymb,mathtools}

\usepackage{graphicx}
\usepackage{adjustbox}
\usepackage{booktabs}
\usepackage{tabularx}
\usepackage{multirow}
\usepackage{xspace}
\usepackage{xcolor}

\usepackage{caption}
\usepackage{subcaption}
\newcommand{\method}{{CrackSegFlow}\xspace}

\usepackage[letterpaper,margin=1.2in]{geometry}  

\usepackage{placeins}
\usepackage{dblfloatfix}
\usepackage{balance}
\usepackage{needspace}

\usepackage[colorlinks=true,linkcolor=black,citecolor=black,urlcolor=blue]{hyperref}

\usepackage[capitalize,noabbrev,nameinlink]{cleveref}

\crefname{figure}{Fig.}{Figs.}
\Crefname{figure}{Fig.}{Figs.}
\crefname{table}{Table}{Tables}
\Crefname{table}{Table}{Tables}
\usepackage[capitalize,noabbrev,nameinlink]{cleveref}

\AtBeginDocument{%
}

\crefname{section}{Section}{Sections}
\Crefname{section}{Section}{Sections}
\crefname{subsection}{Section}{Sections}
\Crefname{subsection}{Section}{Sections}
\crefname{subsubsection}{Section}{Sections}
\Crefname{subsubsection}{Section}{Sections}

\makeatletter
\def\ps@pprintTitle{%
  \let\@oddhead\@empty \let\@evenhead\@empty
  \let\@oddfoot\@empty \let\@evenfoot\@oddfoot}
\makeatother

\journal{Automation in Construction}

\biboptions{numbers,sort&compress}

\begin{document}

\begin{frontmatter}

\title{CrackSegFlow: Controllable Flow Matching Synthesis for Generalizable Crack Segmentation with a 50K Image–Mask Benchmark}

\author[inst1,inst2]{Babak Asadi}

\author[inst1,inst2]{Peiyang Wu}

\author[inst1,inst2]{Mani Golparvar-Fard}

\author[inst1]{Ramez Hajj}

\affiliation[inst1]{organization={Department of Civil and Environmental Engineering, Grainger College of Engineering, University of Illinois Urbana--Champaign},
            city={Urbana},
            state={IL},
            postcode={61801},
            country={USA}}

\affiliation[inst2]{organization={Department of Computer Science, Grainger College of Engineering, University of Illinois Urbana--Champaign},
            city={Urbana},
            state={IL},
            postcode={61801},
            country={USA}}

\begin{abstract}
\itshape
Defect segmentation is central to computer vision–based inspection of infrastructure assets during both construction and operation. However, deployment remains limited due to scarce pixel-level labels and domain shift across environments. We introduce \method, a controllable Flow Matching synthesis method that renders synthetic images of cracks from masks with pixel-level alignment. Our renderer combines topology-preserving mask injection with edge gating to maintain thin-structure continuity. Class-conditional FM samples masks for topology diversity, and CrackSegFlow renders aligned ground truth images from them. We further inject cracks onto crack-free backgrounds to diversify confounders and reduce false positives. Across five datasets and using a CNN–Transformer backbone, our results demonstrate that adding synthesized pairs improves in-domain performance by +5.37 mIoU / +5.13 F1, while target-guided cross-domain synthesis—driven by target mask statistics—adds +13.12 mIoU / +14.82 F1. We also release CSF-50K, a benchmark dataset, comprising 50{,}000 image–mask pairs.

\end{abstract}


\end{frontmatter}

\FloatBarrier
\section{Introduction}
\label{sec:intro}
Automated crack inspection is critical for infrastructure condition assessment and maintenance, 
and it is increasingly used in automated visual inspection workflows (e.g., UAV/robot/vehicle imaging) to support maintenance prioritization and asset-management decisions.
 Advances in computer vision offer accurate, cost-efficient alternatives to manual crack detection and segmentation. Traditional image-processing methods relying on hand-crafted filters or thresholding rules are sensitive to noise and illumination variations \cite{yamaguchi2010fast, fujita2011robust}, while 3D reconstruction approaches can deliver detailed geometric measurements \cite{jahanshahi2012adaptive} but remain resource-intensive to collect and process. As a result, research has shifted toward deep learning (DL) methods that leverage strong feature extraction capabilities to classify \cite{dais2021automatic}, detect \cite{qiu2023real,jiang2024lightweight, mayya2025triple}, and segment \cite{wang2024dual, zhang2025crack, goo2025hybrid, fan2025robotic} cracks. Most notably, pixel-level segmentation directly provides the geometry (length, width, connectivity) needed for severity metrics, eliminating additional post-processing by assigning each pixel as crack or background.  
In Automation in Construction settings, such dense outputs enable consistent, scalable condition assessment and inform maintenance prioritization and asset-management decisions. Therefore, reducing pixel-level labeling cost and improving robustness under domain shift across sites, sensors, and surface textures are key requirements for deployable crack inspection systems.

Existing literature currently proposes segmentation models including CNNs \cite{yamanakkanavar2021novel,lau2020automated, liu2022unet}, Transformers \cite{liu2021crackformer,guo2023novel}, and hybrid CNN\,--\,Transformers \cite{zhou2023hybrid, wang2024dual, goo2025hybrid}. However, the limiting factor is not architectural capacity; it is data scarcity and shift. 
Typical crack datasets contain only a few hundred images collected under specific conditions, with thin/sparse cracks and strong variation in lighting, background texture, crack sparsity, and sensors, which violates the assumption of independent and identically distributed (i.i.d.) variables and degrades cross-domain performance \cite{weng2023unsupervised}. The requirement for pixelwise labels also makes scaling such data particularly expensive.

To address both data scarcity and the high cost of pixel-level annotation, researchers have explored transfer learning \cite{dais2021automatic}, semi-supervised and unsupervised domain adaptation \cite{chun2024self, weng2023unsupervised}, and use of generative models. Early work relied on GANs to synthesize additional crack images \cite{jin2023establishment, pan2023automatic, shim2024self}, which improved performance but often introduced artifacts and required manual annotation of the synthetic images. 
More recently, diffusion models have been applied for realistic crack synthesis and segmentation (including two-stage designs) \cite{jiang2024cracksegdiff, zhang2024vision,han2024crackdiffusion, shi2025crossdiff}. While they can improve quality and accuracy, we observe in practice: (i) iterative sampling increases generation time; (ii) maintaining sub-pixel crack continuity may require additional guidance; (iii) mask-conditioned synthesis tends to mirror the supplied topology; and (iv) two-stage designs can introduce slight mask–image misalignment. Beyond these empirical limitations, latent-diffusion pipelines, such as LDM \cite{rombach2022ldm}, ControlNet \cite{zhang2023adding}, and T2I-Adapter \cite{mou2024t2i} encode mask conditions through concatenation or auxiliary branches within the U-Net’s latent path.  
While effective for general semantic synthesis, these conditioning schemes are sub-optimal for geometry-sensitive structures. The mask signal is gradually weakened by normalization and mixing operations, leading to diluted edge information, loss of sub-pixel continuity, and slight mask–image drift~\cite{lei2025faithful}.

\begin{figure}[ht!]
  \centering
  \includegraphics[width=0.9\linewidth]{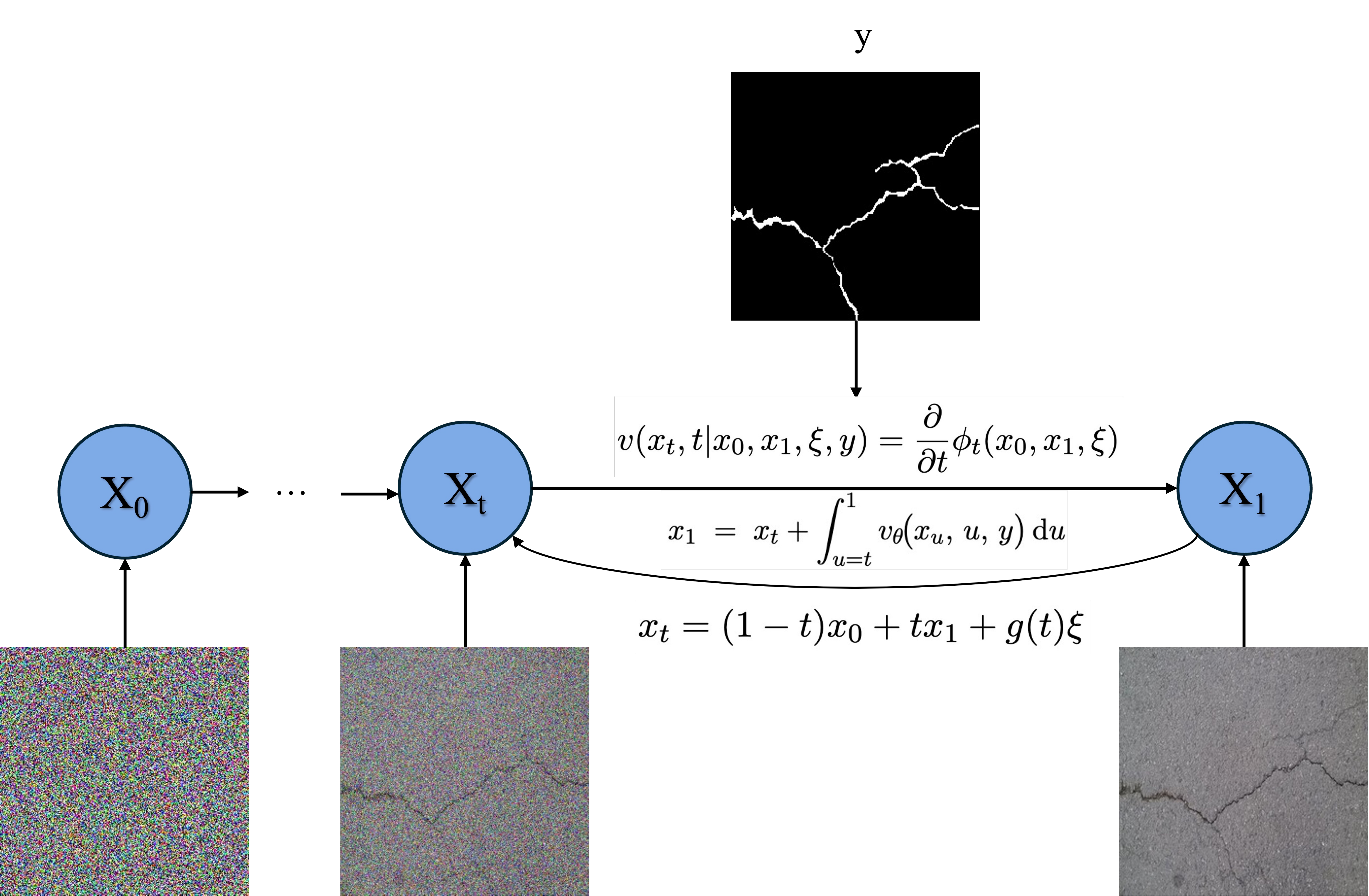}
  \caption{Given mask $y$, sample $x_0\!\sim\!\mathcal N(0,I)$, 
$x_1\!\sim\!p_{\text{data}}(\cdot\mid y)$, $\xi\!\sim\!\mathcal N(0,I)$.
Interpolant $x_t=\phi_t(x_0,x_1,\xi)=(1-t)x_0+tx_1+g(t)\xi$, $g(0)=g(1)=0$.
Oracle velocity $v^*(x_t,t)=\partial_t\phi_t(x_0,x_1,\xi)$; 
train $v_\theta(x_t,t,y)$ to regress $v^*$.
Sampling solves the ODE 
$\frac{\mathrm d x_t}{\mathrm dt}=v_\theta(x_t,t,y)$,
i.e., $x_1=x_0+\int_0^1 v_\theta(x_t,t,y)\,\mathrm dt$.}
  \label{fig:general_process}
\end{figure}

In this work, we introduce \method, a controllable crack-aware generative pipeline built on flow Matching~\cite{lipman2022flow,liu2022flow} (see Fig.~\cref{fig:general_process}). 
At large, \method\ is a two-model FM pipeline: (1) a class-conditional mask generator that discretizes crack coverage into sparsity bins (ultra-sparse$\rightarrow$dense), and (2) a mask-conditioned image renderer that faithfully follows those masks. 
At inference, we sample a target sparsity class to obtain a binary mask, then render a photorealistic image aligned to the topology—yielding annotation-free, topology-controlled pairs. 

The image renderer contains two crack-specific architectural modules. 
The first model is topology-preserving mask injection; a modified SPADE-style normalization applied at every decoder block so the semantic mask persistently modulates features across scales instead of being washed out by normalization. 
The second model is boundary-gated modulation; a lightweight edge gate that selectively amplifies features along crack boundaries, recovering sub-pixel filaments and suppressing texture-driven false positives. 
Together, they act as a geometry-aware attention mechanism that locks synthesis to the input topology and sharpens thin structures. In summary, the major contributions of this work are as follows:

\begin{itemize}
  \item We introduce \method, a controllable Flow Matching (FM) synthesis framework that generates paired crack images and pixel-accurate masks with strict mask--image alignment. The image renderer integrates topology-preserving mask injection and boundary-gated modulation to preserve sub-pixel continuity of thin cracks and suppress texture-driven false positives under deterministic ODE sampling.

  \item We develop a class-conditional FM mask generator that discretizes crack coverage into sparsity bins and supports controllable sampling from ultra-sparse to dense regimes. Combined with lightweight mask propagation, this enables topology-diverse, coverage-balanced supervision and reduces bias toward dataset-specific crack width and annotation conventions.

  \item We propose background-guided crack injection by rendering sampled masks onto crack-free backgrounds, increasing appearance diversity (illumination, shadows, stains, and markings) while keeping background regions labeled as negative. This strategy directly exposes the segmentor to hard negatives such as shadows, joints, and pavement markings and reduces texture-driven false positives.

  \item We validate the proposed synthesis on five benchmarks spanning four asphalt datasets and the crack class of a concrete-domain dataset, under an established hybrid CNN--Transformer segmentation backbone and a fixed training protocol. Augmenting real training data with \method\ pairs improves in-domain performance on average by \mbox{+5.37 mIoU / +5.13 F1} and, with target-guided synthesis using limited target mask statistics, increases the overall cross-domain average by \mbox{+13.12 mIoU / +14.82 F1}.

  \item We release CSF-50K, a public dataset of 50{,}000 sparsity-controlled, topology-diverse image--mask pairs with a 40k/5k/5k train/validation/test split, to support reproducible benchmarking of generalizable crack segmentation.
\end{itemize}

\section{Related Works}
\label{sec:related}

Early supervised approaches to crack segmentation adopted Fully Convolutional Networks (FCNs) \cite{yang2018automatic, dung2019autonomous}, which capture context by progressive downsampling and decode with a single upsampling layer. However, aggressive upsampling can blur small or low-contrast cracks and cause discontinuities \cite{ji2020integrated}. SegNet-based encoder--decoder models \cite{chen2020pavement} improved computational efficiency but can lose local neighborhood information during unpooling of low-resolution feature maps \cite{yamanakkanavar2021novel}. U-Net \cite{ronneberger2015u} advanced the encoder--decoder paradigm with skip connections that preserve fine detail. 
U-Net has therefore become a de facto baseline in crack segmentation, with strong performance reported across multiple datasets \cite{liu2019computer,lau2020automated} and in benchmarks against other CNNs \cite{liu2022unet,tang2023novel,liu2024transfer}. Sensitivity to backbone choice appears limited \cite{liu2022unet}, where training U-Net with VGG-19 \cite{simonyan2014very}, MobileNet-V2 \cite{sandler2018mobilenetv2}, ResNet \cite{he2016deep}, DenseNet \cite{huang2017densely}, and EfficientNet \cite{tan2019efficientnet} as encoder on a 425-image set produced up to a 0.10 mIoU spread (U-Net--VGG19 best), while scaling to a 5330-image composite narrowed the range to 0.01 mIoU \cite{liu2024transfer} with EfficientNet performing best. Comparative studies suggest that DeepLabV3+ can outperform U-Net when very thin cracks are not predominant \cite{ji2020integrated, shamsabadi2022vision}, while its dilated convolutions may still overlook sub-pixel structures \cite{shamsabadi2022vision}.

To address the limited locality of convolutions, Transformer-based crack segmentors have been explored to capture long-range dependencies \cite{liu2021crackformer, wang2022automatic, guo2023novel}. The SegCrack model \cite{wang2022automatic}, a hierarchical Transformer encoder with a top-down decoder and lateral connections, outperformed several models but was surpassed by DeepLabV3+ on another dataset in \cite{zhou2023hybrid}. Guo et al. \cite{guo2023novel} combined a Swin Transformer \cite{liu2021swin} encoder with UPerNet and attention in the decoder, achieving F1 scores of 0.842 and 0.764 for two datasets, which was marginally higher than U-Net (0.837 and 0.749). 
Li et al. \cite{li2024automatic} evaluated SegFormer \cite{xie2021segformer} across public datasets and reported an overall mIoU of 0.778, merely 0.007 above a tuned U-Net. 
In sub-dataset comparisons, U-Net--VGG16 surpassed SegFormer on CrackTree and DeepCrack, which differ from each other in terms of crack-pixel prevalence. Hybrid CNN--Transformer designs aim to retain local detail while modeling global context \cite{shamsabadi2022vision, zhou2023hybrid, quan2023crackvit, xiang2023crack, wang2024dual, goo2025hybrid}. For instance, Zhou et al. \cite{zhou2023hybrid} integrated a convolutional inverse residual with a Swin block in the encoder, improving mIoU by 0.03 over DeepLabV3+.

The literature has largely centered on architectural innovation evaluated on a handful of public datasets, with classical CNNs occasionally performing similarly to leading Transformer models. 
However, fully supervised models, irrespective of architectural design, struggle to generalize and adapt under domain shift across datasets. For example, the dual-path CNN--Transformer in \cite{goo2025hybrid} achieved mIoU of 0.87 (DeepCrack \cite{liu2019deepcrack}) and 0.804 (CrackTree \cite{zou2018deepcrack}) in-domain, but dropped to 0.653 (DeepCrack$\rightarrow$CrackTree) and 0.585 (CrackTree$\rightarrow$DeepCrack) in cross-domain evaluations. 
Similar trends were observed by Quan et al. \cite{quan2023crackvit}. Thus, it can be inferred that deep learning models for crack segmentation are highly specialized to their training datasets. 
Furthermore, adapting these models to each new dataset requires annotating hundreds of images, a process that is labor-intensive and impractical.

While transfer learning and domain adaptation can mitigate distribution shift, generative augmentation directly expands training data with controllable geometry and appearance and can produce paired image--mask data without additional annotation at inference. This capability is particularly valuable for thin/sparse cracks. 
Consequently, we focus on generative methods. 
We first apply GANs and diffusion, which have both previously been applied for crack segmentation.
Subsequently, we introduce the application of flow Matching, a deterministic alternative well suited to preserving fine detail and topology, which forms the foundation of the model developed in this work. Early works used GANs to synthesize crack imagery or paired examples. Chen et al. \cite{chen2022crack} expanded training data fivefold with DCGAN and found that despite artifacts and the need to annotate synthetic samples, retraining DeepLabV3+ improved performance. Jin et al. \cite{jin2023establishment} generated pseudo-annotations via DCGAN and rendered images via Pix2Pix, achieving 74.34\% of the mIoU of a real-data model, with further gains after mixing synthetic and real images. CrackSegAN \cite{pan2023automatic} similarly exhibited slight improvements. While impactful, GAN synthesis can introduce texture artifacts, exhibits limited topology diversity (mode collapse), and often requires manual curation or labeling to ensure usable pairs---constraints that limit scalability when segmenting thin and sparse cracks.

Diffusion models have become the state-of-the-art for high-fidelity visual synthesis. 
Denoising diffusion probabilistic models (DDPM) sample by iterative denoising \cite{ho2020ddpm}, classifier-free guidance (CFG) improves conditional fidelity without an external classifier \cite{ho2022classifierfree}, 
and latent diffusion reduces runtime by operating in a learned latent space \cite{rombach2022ldm}. Within crack analysis, diffusion appears in two principal roles. 
For augmentation, mask-conditioned or semantics-guided diffusion renders crack images from supplied layouts to enrich training datasets. 
For example, \cite{zhang2024vision} employed a denoising-diffusion refiner (CrackSegRefiner) to improve pixel-level crack masks for a vision-guided sealing robot. 
\cite{jiang2024cracksegdiff} proposed a diffusion-based segmenter that fuses grayscale and depth. 
Second, diffusion is important as part of two-stage pipelines. \cite{han2024crackdiffusion} mapped crack-containing images toward a crack-free distribution in an unsupervised diffusion stage and concatenated the result with the original image to boost a subsequent U-Net. 
\cite{shi2025crossdiff} introduced a cross-conditional diffusion segmentor aimed at better thin-structure retention. 
Overall, diffusion-based augmentation indicates that high-quality crack synthesis is feasible and can strengthen in-domain accuracy and, in some cases, cross-domain robustness. 
At the same time, there are several drawbacks. Iterative sampling can raise generation time relative to one-shot generators. Preserving sub-pixel continuity may benefit from additional guidance. Mask-conditioned rendering tends to mirror the supplied topology, limiting diversity if masks have narrow distributions. Finally, two-stage designs can introduce slight mask--image misalignment. 
These considerations motivate generative frameworks that maintain fidelity to fine structures while enabling controllable, efficient pair synthesis.

Flow Matching (FM) \cite{lipman2022flow} and Rectified Flow \cite{liu2022flow} generate by learning a time-dependent vector field that transports a simple base distribution to data along a prescribed path, enabling deterministic ODE sampling. Training regresses closed-form pairwise velocities along the path, which yields stable objectives and few-step sampling. In dense prediction, two directions have emerged. First, unified Rectified Flow models bind image and mask spaces so a single network can both synthesize images and predict segmentation maps \cite{wang2024semflow,caetano2025symmflow}. Second, segmentor-centric FM variants predict masks from images using flow-based objectives \cite{bogensperger2025flowsdf}. These works demonstrate FM's viability but differ from an external, controllable augmentation engine; prior flow-based designs typically do not parameterize crack topology (e.g., explicit sparsity classes) nor guarantee geometry-consistent image rendering from masks at inference without additional labels.

Given that architectural gains are modest using small datasets and that domain shift remains the key obstacle, generative augmentation is a natural lever. GANs help, but provide limited controllability over topology while diffusion provides high-fidelity synthesis and paired data but at higher sampling cost and with limited control over mask diversity when conditioning sources are narrow. FM/Rectified flow offers deterministic, few-step sampling and strong crack fidelity. These properties motivate the design in the next section: a two-model, controllable FM pipeline comprising (i) a class-conditional FM mask generator that sweeps crack sparsity and (ii) a mask-conditioned FM image generator that renders geometry-consistent images from those masks.  This pipeline enables annotation-free pair creation at inference and targeting cross-domain robustness.

\section{Methodology}
We propose a crack-aware generative framework built on Conditional Flow Matching (FM) for semantic image synthesis, coupled with a second FM model that generates crack masks conditioned on coarse sparsity classes. Each class corresponds to a bin of crack-pixel coverage (percentage of crack pixels in the image); for example, if crack coverage spans 0--5\%, we define classes such as 0--0.5\% (class~0), 0.5--1\% (class~1), \ldots, 4.5--5\% (class~9). In contrast to denoising diffusion pipelines, our training target is an exact velocity field derived from an analytic interpolation between a simple base distribution and the data, as shown in ~\cref{fig:general_process}. Sampling is deterministic (ODE integration) and does not re-inject noise, which helps preserve hairline structures typical of crack imagery. The architectural backbone remains a SPADE-conditioned U-Net so that improvements can be attributed to the FM formulation and our conditioning design rather than to capacity changes. This section presents (i) the motivation and problem setting, (ii) preliminaries on the conditional FM formulation and its learning objective, (iii) a detailed description of the proposed CrackSegFlow, (iv) the class-conditional mask generator and its integration during inference, and (v) mask injection into crack-free backgrounds via rectified flow, which increases variation in background illumination and texture and thereby reduces false positives caused by background patterns.

\subsection{Problem Setting and Motivation for Flow Matching}
\begin{figure*}[t]
  \centering
  \includegraphics[width=0.95\textwidth]{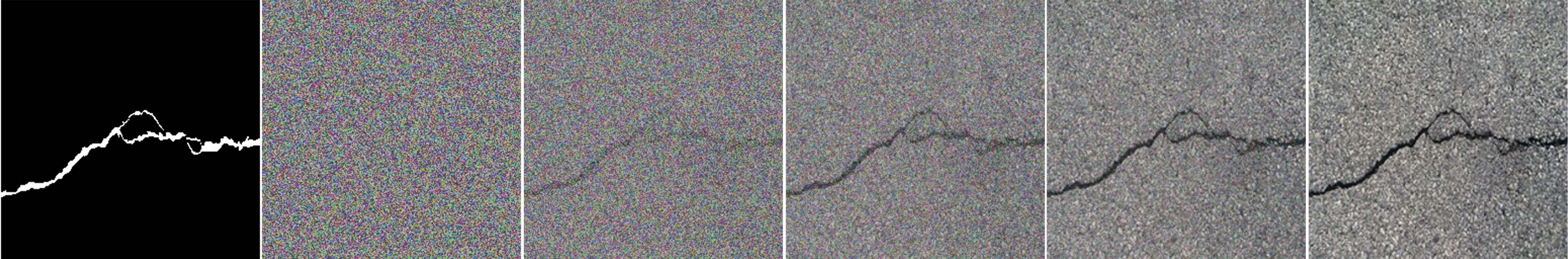}
  \caption{Noise to Image with Crack.}
  \label{fig:noise_to_image}
\end{figure*}

Let \(x \in \mathbb{R}^{H\times W\times C}\) denote a crack image and \(y \in \{0,1\}^{H\times W}\) a binary crack mask. As in ~\cref{fig:noise_to_image}, the goal is to learn a conditional generator that samples \(x\) faithful to \(y\) while maintaining photorealism and diversity. Crack pixels often occupy \(<3\%\) of the image, and their geometry consists of thin, high-curvature filaments and junctions. Methods that rely on iterative stochastic denoising can blur or detach these structures unless many steps and carefully tuned guidance are applied. Our design replaces stochasticity with a learned velocity field \(v_\theta(x,t,y)\) that transports samples along a continuous path \(\{p_t(\cdot\mid y)\}_{t\in[0,1]}\) from a base \(p_0\) to the data \(p_1(\cdot\mid y)\). This design has two practical consequences. First, supervision is low-variance and directly tied to the data via a closed-form target, so the model is forced to explain pixelwise transport including within sparse crack regions. Second, the deterministic sampler avoids repeated noise perturbations, stabilizing thin structures with far fewer integration steps. To systematically probe robustness to domain shift and sparsity, we complement the image model with a mask-only FM generator conditioned on discretized crack coverage; the latter is used only at inference to produce a balanced set of masks.

\subsection{Preliminaries}

Flow Matching (FM) treats generation as transporting a simple base distribution \(p_0\) (e.g., \(\mathcal N(0,I)\)) to the data distribution by learning a time-dependent velocity field that evolves a path of densities \(\{p_t\}_{t\in[0,1]}\) (optionally conditioned on a label or mask). The evolution is governed by the continuity equation:

\begin{equation}
\partial_t p_t(x\mid y)\;+\;\nabla_x\!\cdot\!\big(v_\theta(x,t,y)\,p_t(x\mid y)\big)\;=\;0.
\label{eq:pre_eq1}
\end{equation}

At inference, a sample is obtained by solving the deterministic ODE (probability-flow dynamics) from an initial draw \(x_0\sim p_0\):
\begin{equation}
\frac{d}{dt}\,x_t\;=\;v_\theta(x_t,t,y),\qquad t\in[0,1],\qquad x_1\approx p_{\text{data}}(\cdot\mid y).
\label{eq:pre_eq2}
\end{equation}

Learning \(v_\theta\) avoids simulating (2) during training by supervising it with pairwise velocities along an analytically specified path between base and data. Let \((x_0,x_1)\sim p_0\times p_{\text{data}}(\cdot\mid y)\) and let \(\xi\sim\mathcal N(0,I)\). An interpolant \(\varphi_t\) maps the pair (and optionally noise) to an intermediate state:

\begin{equation}\label{eq:pre_eq3}
\begin{aligned}
x_t &= \varphi_t(x_0,x_1,\xi)
     = \alpha(t)\,x_0 + \beta(t)\,x_1 + g(t)\,\xi,\\
&\qquad \alpha(0)=1,\ \beta(1)=1,\ g(0)=g(1)=0.
\end{aligned}
\end{equation}

which includes common choices such as linear displacement (\(\alpha{=}1{-}t,\ \beta{=}t,\ g{\equiv}0\)) or stochastic interpolants (nonzero \(g\) that vanishes at the endpoints). The target (pairwise) velocity is the time derivative of the interpolant:
\begin{equation}
u_t(x_0,x_1,\xi)\;=\;\partial_t\,\varphi_t(x_0,x_1,\xi)
\;=\;\dot\alpha(t)\,x_0\;+\;\dot\beta(t)\,x_1\;+\;\dot g(t)\,\xi.
\label{eq:pre_eq4}
\end{equation}

FM trains the model by regressing the predicted velocity to this target at randomly sampled times \(t\sim\mathcal U[0,1]\), yielding a low-variance, closed-form objective:
\begin{equation}
\mathcal L_{\text{FM}}
\;=\;
\mathbb E_{\substack{x_0,x_1,\xi,\,t\\ y}}
\Big[\ \big\|\,v_\theta\big(\varphi_t(x_0,x_1,\xi),\,t,\,y\big)\;-\;u_t(x_0,x_1,\xi)\,\big\|_2^2\ \Big].
\label{eq:pre_eq5}
\end{equation}

Intuitively, (5) teaches the network to point from the current interpolated state \(x_t\) toward the data endpoint \(x_1\) (and away from \(x_0\)) in a manner consistent with the chosen path (3). Once trained, integrating (2) deterministically transports \(p_0\) to an approximation of \(p_{\text{data}}(\cdot\mid y)\) in few ODE steps, avoiding stochastic denoising and preserving thin structures.

\begin{figure*}[t]
  \centering
  \includegraphics[width=0.99\textwidth]{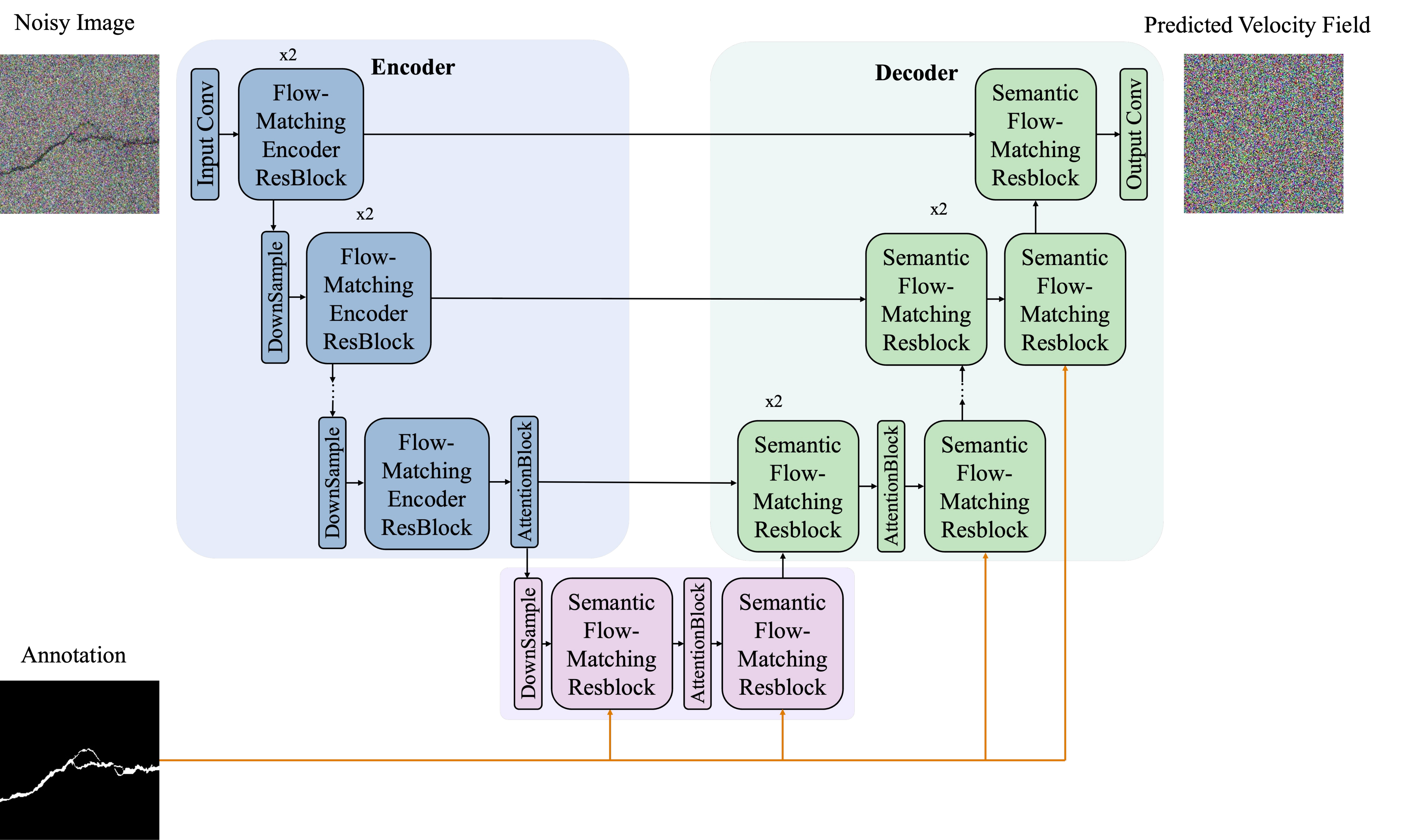}
  \caption{U\!-Net backbone used in our  branch.}
  \label{fig:unet}
\end{figure*}

\begin{figure}[h!]
  \centering
  \includegraphics[width=0.85\linewidth]{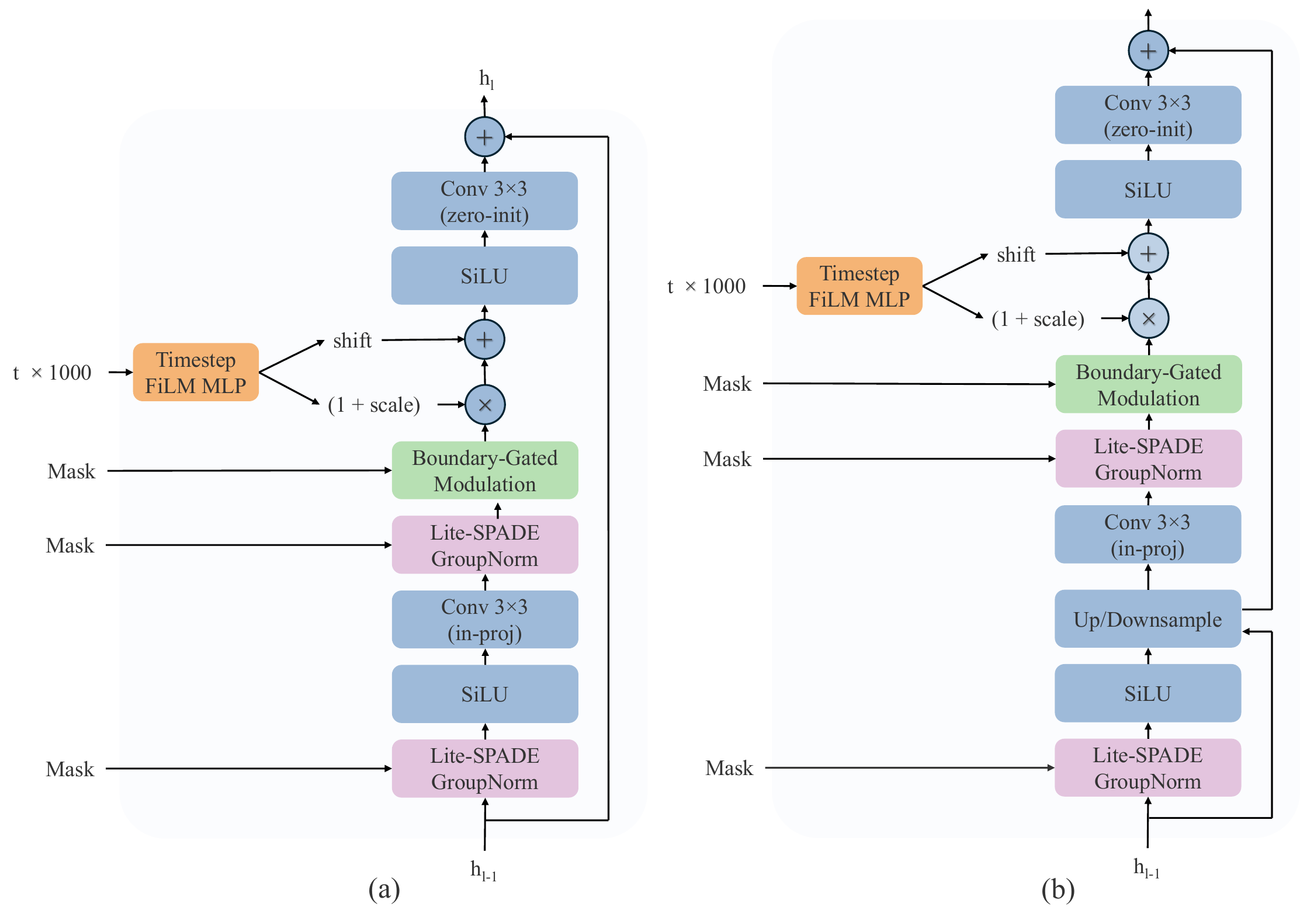}
  \caption{CrackSegFlow residual block used in the velocity-field network $v_\theta$.}
  \label{fig:sfmresblock}
\end{figure}

\begin{figure}[h!]
  \centering
  \includegraphics[width=0.8\linewidth]{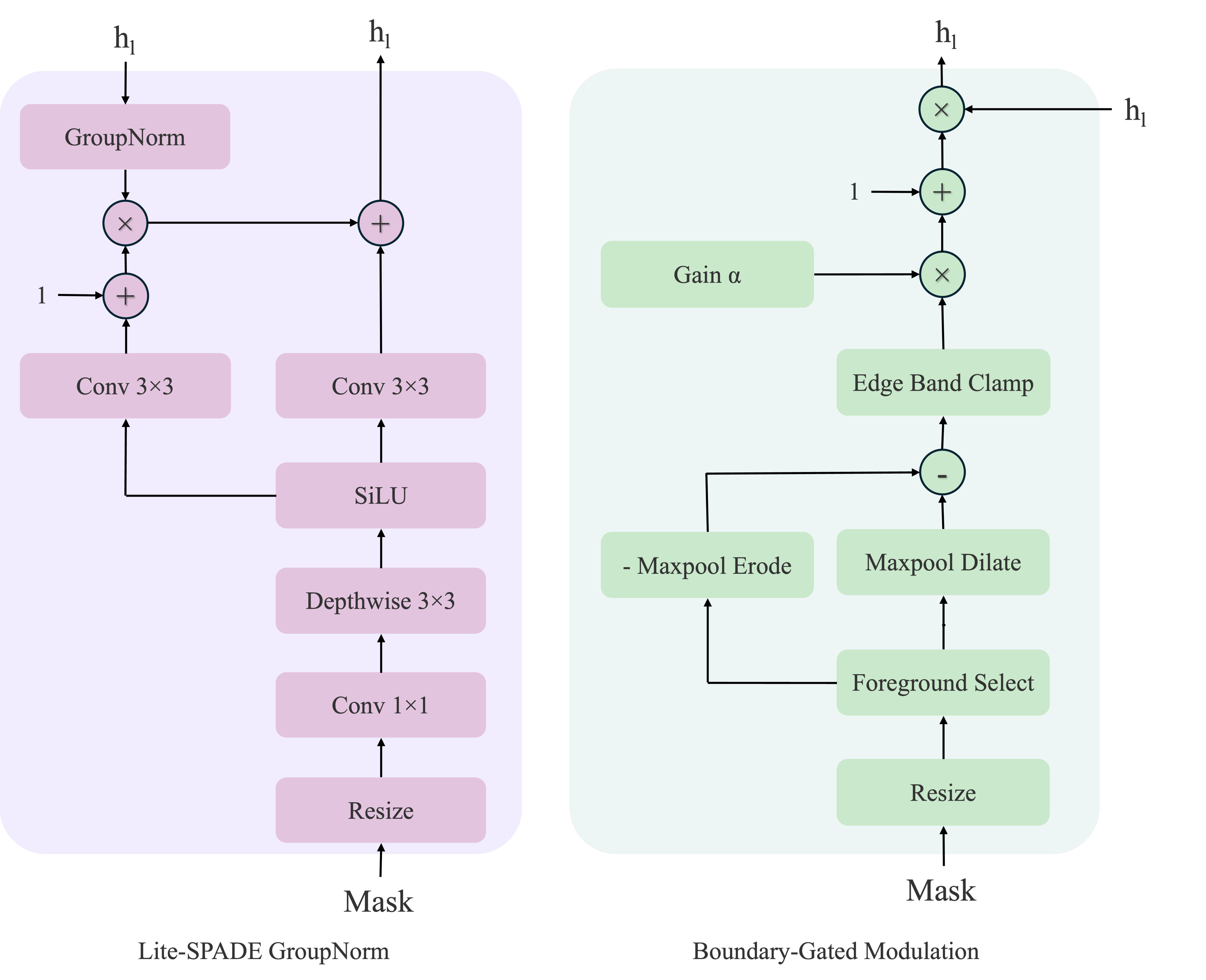}       
  \caption{Topology-Preserving Mask Injection paired with Boundary-Gated Modulation. The former is our modified SPADE-style conditioning applied at every decoder block to preserve mask topology across scales, while the latter selectively amplifies features near crack boundaries to recover sub-pixel filaments.}
  \label{fig:spade-edge}
\end{figure}

\subsection{Semantic Flow Matching Model}
CrackSegFlow is built on a modified U-Net architecture with an encoder–decoder symmetry and strong skip connections at each resolution. As shown in detail in \cref{fig:unet}, we adopt a U-Net-style backbone with residual blocks and multi-scale attention, closely following widely used diffusion backbones but without structural changes.

\subsubsection{Basic Settings of U-Net}
The encoder progressively downsamples the input (starting from a noise image) through convolutional layers, and the decoder symmetrically upsamples it back to full resolution. Skip connections link each encoder layer to its counterpart in the decoder to ensure that high-frequency details are carried forward. We include multi-head self-attention blocks \cite{wang2018non} with skip connections, which are formulated as follows:

\begin{equation}\label{eq:attention}
\begin{aligned}
f(x)&=\mathcal{W}_f x,\quad g(x)=\mathcal{W}_g x,\quad h(x)=\mathcal{W}_h x,\\
\mathcal{M}(u,v)&=\frac{f(x_u)^{\!\top} g(x_v)}{\lVert f(x_u)\rVert\,\lVert g(x_v)\rVert},\\
y_u&=x_u+\mathcal{W}_v\sum_{v'} \operatorname{softmax}_{v'}\!\bigl(\alpha\,\mathcal{M}(u,v')\bigr)\,h(x_{v'}).
\end{aligned}
\end{equation}
where $\alpha$ is a learnable temperature controlling attention sharpness.

In these equations,  $x$ and $y$ denote the input and output of the attention layer, and $\mathcal{W}_f$, $\mathcal{W}_g$, $\mathcal{W}_h$, and $\mathcal{W}_v \in \mathbb{R}^{c \times c}$ refer to $1 \times 1$ convolutional operators in the attention layer. $u$ and $v$ index spatial locations ranging from 1 to $H \times W$. In our structure, we attach attention blocks to ResBlocks at some specific resolutions, i.e., 32 \(\times\) 32, 16 \(\times\) 16, 8 \(\times\) 8, to let the network attend across distant regions. 

We also pass the timestep to make the network aware of the current position for velocity estimation. Both encoder and decoder have learnable $\omega(t), b(t) \in \mathbb{R}^{1 \times 1 \times C}$ that scale and shift features. We scale $t$ by $10^3$ before the time embedding:
\begin{equation}
f^{i+1}=\omega(10^3 t)\cdot f^{i} + b(10^3 t)
\label{eq:SFM_timestep}
\end{equation}
where $f^{i}$ and $f^{i+1}$ are the input and output features of layer $i$.

\subsubsection{Topology-Preserving Mask Injection}\label{sec:tpmi}

We inject the semantic mask into the U-Net to guide the velocity field generation. In diffusion pipelines, previous work directly concatenates the semantic mask with the noisy image as input and guides the conditioning \cite{saharia2022palette, saharia2022image}. It has been shown that this can fail to carry the semantic information because standard normalization tends to attenuate layout signals; injecting masks via spatially-adaptive normalization preserves the spatial semantics during decoding \cite{park2019semantic, wang2022semantic}. 

To resolve this issue, we designed a CrackSegFlow residual block (CrackSegFlowResBlock) constructed with Spatially-Adaptive Denormalization (SPADE) layers to persistently inject the binary crack mask at every decoder block (see \cref{fig:sfmresblock}). SPADE is a conditional normalization technique that preserves spatial layout information by modulating feature activations with learned scale and bias maps derived from input segmentation \cite{park2019semantic}. Concretely, we adopt SPADE \cite{park2019semantic} as the base operator but (i) apply it at all decoder normalizations, (ii) use a lightweight per-scale mask encoder to produce $\gamma_l(M)$ and $\beta_l(M)$, and (iii) couple it with the edge-focused gate in \cref{sec:bgm}. In our context, the binary crack mask (where crack pixels are labeled and background is unlabeled) is used to modulate the decoder feature maps so that the network knows where cracks are supposed to appear. In processing, let $h^l$ denote the decoder feature at level $l$. We first apply a normalizing transform to $h^l$ (we use group normalization for stability), yielding $h^l_{\text{norm}}$; let $M$ be the mask resized to that level. A small convolutional network then takes $M$ as input and produces $\gamma_l(M)$ and $\beta_l(M)$ of equal dimension to $h^l_{\text{norm}}$, keeping the mask signal explicit within the normalization path. The feature is then modulated as:
\begin{equation}
h^{l}_{\text{out}}(x) = \gamma_l(M)_x \cdot h^{l}_{\text{norm}}(x) + \beta_l(M)_x
\label{eq:SFM_SPADE}
\end{equation}
where $x$ indicates spatial location (pixel). The CrackSegFlowResBlock essentially performs a per-pixel affine transformation on the feature map, with coefficients that depend on the local semantic label (crack or background). With this structure, the crack mask “injects” information by altering feature activations differently at crack pixels versus background pixels. We also adapted an upsample/downsample layer in the CrackSegFlowResBlock (see \cref{fig:sfmresblock}(b)), which changes the resolution of current features for passing through different levels of the U-Net. By utilizing the CrackSegFlowResBlock in multiple decoder blocks, we make the semantic mask influence generation at every scale. The network cannot ignore the mask since the semantic layout is reinforced after each normalization, which prevents the common problem of segmentation information being washed out through normalizations. 

Our semantic mask injection process is shown in \cref{fig:spade-edge}(a), using a lightweight mask encoder within each SPADE, which we refer to as Lite-SPADE GroupNorm. The main part is a shared $3\times3$ conv $+$ ReLU that projects the mask (2 channels for foreground and background) to an intermediate embedding, which is then converted to the $\gamma$ and $\beta$ maps via separate $3\times3$ conv layers. It should be noted that GroupNorm is applied to features only (separate from the semantic masks), so the information of masks is preserved. We found that injecting the mask at every CrackSegFlowResBlock in the decoder (both the first normalization and the second normalization in each block) yielded good detail preservation. This approach aligns with recent diffusion-based synthesis models, which feed the noisy image through the encoder and the semantic layout through multi-layer adaptive normalization in the decoder, resulting in improved precision to the input layout. 

\subsubsection{Boundary-Gated Modulation}\label{sec:bgm}

We pair topology-preserving mask injection with a lightweight boundary-gated modulation that boosts features along crack edges. While SPADE injects the mask globally, we noticed that the semantic mask has an imbalanced distribution. As shown in \cref{tab:datasets}, crack proportion ranges from $0.36\%$ to $2.84\%$. Cracks in the mask are often only 1–2 pixels wide, so without special handling, their features could be diluted by surrounding background features. To address this issue, we introduce a novel boundary-gated godulation to focus the network’s capacity on the thin, minor crack pixels, shown in \cref{fig:spade-edge}(b). This module addresses this by selectively enhancing feature responses around crack boundaries. It computes a boundary confidence map $G(x)$ from the crack mask, highlighting the immediate vicinity of crack edges. We implement $G$ as a morphological gradient (dilation minus erosion) of the mask, producing a binary edge map; optionally we thicken edges and normalize to $[0,1]$ if a wider band is desired. The gating map $G(x)$ is 1 at crack edge pixels and 0 elsewhere (or decayed between 0–1 if a thicker border is considered). The feature activations after SPADE are then modulated using a gating function. Let $h_{\text{spade}}(x)$ be the feature after SPADE normalization and $G(x)$ the boundary map (we treat background as 0 and crack boundary as 1). We introduce a learnable scalar parameter $\omega$ and define the gated output as
\begin{equation}
h_{\text{gated}}(x) \;=\; h_{\text{spade}}(x)\,\bigl(1 + \omega\,G(x)\bigr),
\label{eq:SFM_gate}
\end{equation}
where $G(x)$ is broadcast across channels and spatially aligns with $h_{\text{spade}}(x)$.

This can be seen as scaling the SPADE affine parameters by $(1 + \omega,G(x))$, i.e. $\gamma'(x) = \gamma(x)(1 + \omega, G(x))$ and $\beta'(x) = \beta(x)(1 + \omega,G(x))$. When $G(x)=1$ (on a crack boundary), the feature activation is amplified by $(1+\omega)$; when $G(x)=0$ (away from cracks), the feature passes through unchanged. The learned gain $\omega$ allows the network to adjust how strongly to boost crack-edge features. In practice, we apply this boundary gate in each SPADE-modulated ResBlock of the decoder, right after the SPADE normalization. Starting with $\omega=0$ ensures that initially there is no difference, and during training, the model can increase $\omega$ if emphasizing edges improves the reconstruction of thin cracks. This gating strategy is lightweight—using a morphological edge map and a single scalar per layer. But it effectively highlights crack pixels in the feature maps so they are not overshadowed by surrounding background features. 

Following the general routine of generative models, it is noticeable that they might not be strongly correlated with conditional labels. In our implementation, the potential risk that the model doesn't precisely follow the semantic masks will result in failure if a segmentation model is trained on it. 

It is worth mentioning that, to improve conditional fidelity without introducing an auxiliary classifier, we adopt classifier-free guidance (CFG)~\cite{ho2022classifierfree} during training and sampling. With probability $p_{\mathrm{drop}}$, we replace the crack mask $y$ with an empty mask $y_{\emptyset}$ when computing the loss.

\begin{equation}
    \hat{v}_{\theta}(x, t, y) = v_\theta(x, t, y_\emptyset) + \omega (v_\theta(x, t, y) - v_\theta(x, t, y_\emptyset)) 
\end{equation}

\noindent At inference, we solve~\(\displaystyle x_1 = x_0 + \int_0^1 \hat v_\theta(x_t,t,y)\,dt\), yielding a synthetic image \(x_1\). The gain \(\omega\) trades diversity for adherence to the requested coverage class.

\subsection{Semantic Mask Synthesis}

In addition to the CrackSegFlow, we also train a conditional FM model that synthesizes crack masks directly. Previous work has shown semantic-synthesis pipelines by reusing the original masks. Our mask generator learns from the distribution of a binary mask dataset with precise control of crack proportion. For each mask with $M \in \{0, 1\}^{(H \times W)}$, we label it with a discrete control label $y$ based on the fraction of crack pixels:

\begin{equation}
\rho(M) = \frac{1}{HW} \sum_{i,j} M_{ij}, \quad
y = \mathcal{B}(\rho(M)) \in \{0,\dots,C-1\}
\label{eq:mask_label}
\end{equation}

\noindent where $\mathcal{B}$ bins $\rho$ into $C$ intervals, from very small to very large cracks. At inference, sampling with a target y gives masks with the desired proportion, paired image–mask generation when combined with our CrackSegFlow.

The mask generator uses a U-Net structure but with 1-channel in/out, with residual blocks, skip connections, and multi-scale self-attention at lower resolutions and predicts the instantaneous velocity field $v_\theta(x, t, y)$. The conditioning on y is implemented as a label embedding that augments the standard sinusoidal time embedding. Let $\psi(t) \in \mathbb{R}^d$ denote the time embedding and let $E:\{0,\dots, C-1\} \rightarrow \mathbb{R}^d$ be a trainable embedding parameter. The model forms:


\begin{equation}
z=\phi(\psi(t) + E(y)) \in \mathbb{R}^{d}
\label{eq:mask_embedding}
\end{equation}

\noindent where $\phi$ is a small MLP. In each ResBlock, $z$ drives feature modulation through the network, and it is mapped and added to the block’s hidden activation. By injecting the embedding $z$ to every block throughout the U-Net, we control the proportion of cracks in each synthetic semantic mask.

Like CrackSegFlow, the semantic mask generator also utilized classifier-free guidance for conditional fidelity. One small difference is that y is replaced with a null value $\emptyset$, which is omitted from the embedding process, and adjusted velocity field is: $\hat{v}_{\theta}(x, t, y) = v_\theta(x, t, \emptyset) + \omega (v_\theta(x, t, y) - v_\theta(x, t, \emptyset))$ which is used for ODE integration.

\begin{figure}[t]
  \centering
  \includegraphics[width=0.5\linewidth]{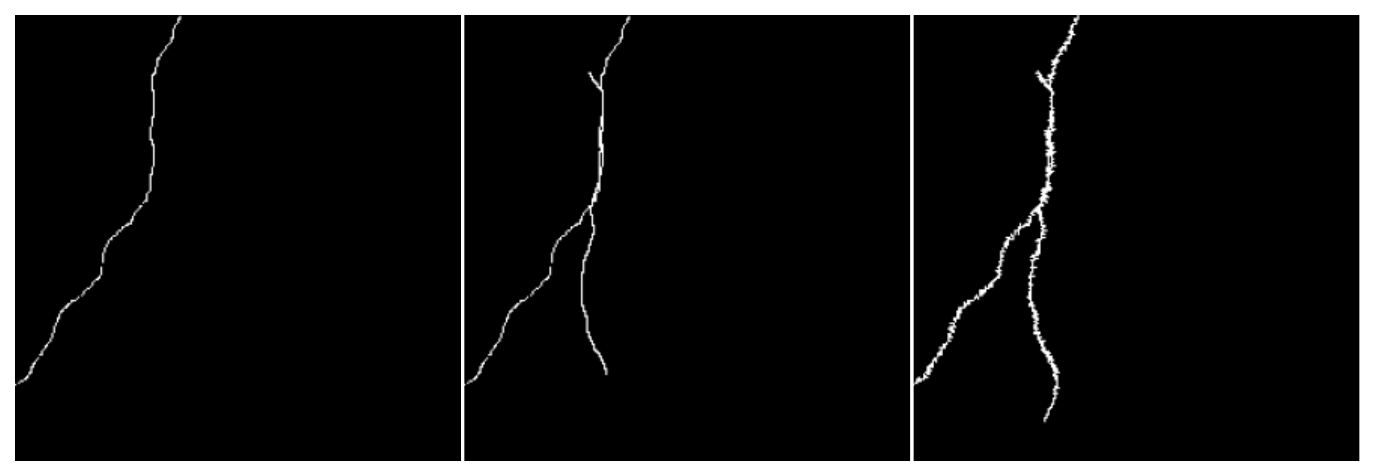}
  \caption{Example of mask propagation. Starting from the same base crack mask, we generate multiple structure-preserving propagated variants to broaden intra-regime crack geometry.}
  \label{fig:mask_propagation}
\end{figure}

Beyond class-conditional sampling, we further expand topology diversity via mask propagation. While sparsity bins control the global crack-pixel ratio, they do not explicitly enforce local geometric variations such as centerline perturbations, small branching, or width changes that arise across sensors and annotation styles. Therefore, given a base mask $M$, we generate a small family of propagated masks $\{M^{(j)}\}_{j=0}^{K-1}$ by applying lightweight, structure-preserving morphological perturbations (e.g., controlled dilation/erosion, local thinning/thickening, and small spatial jitters) while maintaining connectivity and avoiding topological collapse. In practice, for each training mask we save multiple propagated variants (e.g., $K{=}3$), and use them as additional conditioning inputs to the image renderer. This augmentation is label-preserving (all variants remain binary masks) and complements the coverage-based control: sparsity conditioning selects the regime, while propagation broadens intra-regime geometry. Empirically, this improves robustness to annotation conventions (1-pixel centerlines versus thicker traces) and reduces overfitting to dataset-specific crack morphology. An example of these propagated variants is shown in ~\cref{fig:mask_propagation}, where three morphologically perturbed versions of the same crack mask are visualized.

\subsection{Rectified Flow for Background-Guided Crack Injection}
Crack segmentation models often fail not because cracks are absent, but because backgrounds contain crack-like patterns such as shadows, joints, stains, and pavement markings. Standard mask-conditioned synthesis improves crack topology and appearance, yet it may not sufficiently expose the segmentor to these hard negatives under diverse, crack-free contexts. To directly increase background diversity while keeping pixel-accurate supervision, we propose background-guided crack injection: given any crack-free background, we render a realistic crack instance that strictly follows a provided mask, producing aligned image--mask pairs where all non-masked regions remain negative.

\begin{figure}[!t]
  \centering
\includegraphics[width=0.8\linewidth,keepaspectratio]{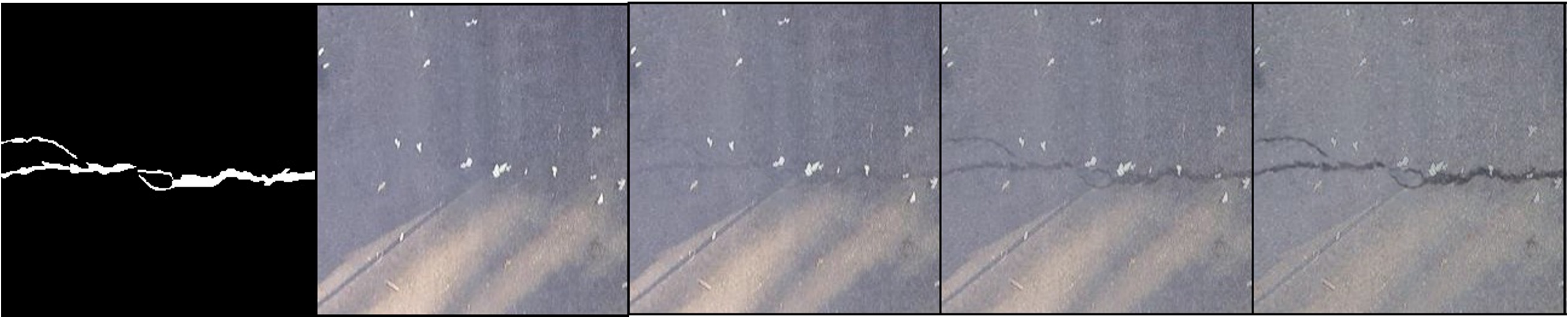}
  \caption{Inject a given crack on any background by rectified-flow.}
  \label{fig:background_to_image}
\end{figure}

With a desired crack mask $M\in{0,1}^{H\times W}$ and an arbitrary background image $x_0\in\mathbb{R}^{H\times W\times 3}$, we learn a conditional, time-indexed velocity field $v_\theta(x,t,M)$ that deterministically transports $x_0$ to a target crack image $x_1\sim p_{\text{data}}(\cdot\mid M)$ along a probability-flow ODE (see \cref{fig:background_to_image}).

Training (rectified flow Matching). We draw $t\sim\mathcal{U}[0,1]$ and use a rectifying schedule $\phi(t)=t^2$ with $\phi'(t)=2t$. The stochastic interpolant and its closed-form pairwise velocity are:

\begin{equation}
x_t=(1-\phi(t))\,x_0+\phi(t)\,x_1+\sigma\,\sqrt{\phi(t)\,(1-\phi(t))}\,\varepsilon.
\label{eq:rf_xt}
\end{equation}

\noindent with $\varepsilon\sim\mathcal{N}(0,I)$, and $,u_t=\phi'(t),(x_1-x_0),$. We regress the network to $u_t$ via:

\begin{equation}
\mathcal{L}_{\mathrm{RF}}
=\mathbb{E}\Big[\big\|v_\theta(x_t,t,M)-u_t\big\|_2^2\Big].
\label{eq:rf_loss}
\end{equation}

The mask is provided as a per-pixel one-hot map $y$ (two classes), with mild condition dropout during training. Each step samples $(x_1,y)$ from the crack dataset and an independent background $x_0$ from a background loader, forms $(x_t,u_t)$, and minimizes $,|v\theta(x_t,t,y)-u_t|_2^2,$.

Given a new background $b$ and mask $M$, we solve the deterministic ODE
$,\frac{dx}{d\tau}=v_\theta(x,\tau,M),\quad x(0)=b,\ \tau\in[0,1],$
with a fixed-step Euler integrator (e.g., $K$ uniform steps). The terminal state $x(1)$ preserves the background context while injecting the requested crack topology specified by $M$. This complements our standard FM formulation by using the rectified schedule to “straighten” transport from background to target, yielding noise-free, deterministic sampling.

\section{Experiments}

\subsection{Datasets}
\label{subsec:datasets}

In this study, five datasets were adopted to evaluate the proposed framework.
CrackTree260 \cite{zou2012cracktree, zou2018deepcrack} includes pavement images acquired under visible-light illumination using an area-array camera.
Crack500 \cite{zhang2016road} comprises high-resolution crack images collected under diverse lighting and texture conditions using handheld mobile phones. CrackLS315 \cite{zou2018deepcrack} contains asphalt pavement images captured by a line-array camera under laser illumination.
CFD \cite{shi2016automatic} contains asphalt pavement crack images with varied surface textures and imaging conditions.
In addition, we include the crack class from the Structural Defects Dataset (S2DS) \cite{benz2022image}, which features concrete surfaces and introduces a distinct material domain relative to the asphalt-focused datasets above.
A summary of dataset characteristics is provided in ~\cref{tab:datasets}.

\begin{table}[t]
\centering
\caption{Summary of benchmark crack datasets used in this study.}
\label{tab:datasets}
\setlength{\tabcolsep}{8pt}
\renewcommand{\arraystretch}{1.05}
{\footnotesize
\begin{adjustbox}{max width=\columnwidth}
\begin{tabular}{lccc}
\toprule
\textbf{Dataset} & \textbf{Resolution} & \textbf{\# Images} & \textbf{Crack Proportion (\%)} \\
\midrule
CrackTree260 & [600, 720] $\times$ [800, 960] & 260 & 0.46 \\
CRACK500     & [1440, 1936] $\times$ [2560, 2592] & 500 & 2.84 \\
CrackLS315   & 512 $\times$ 512 & 315 & 0.25 \\
CFD          & 320 $\times$ 480 & 118 & 1.62 \\
S2DS         & 1024 $\times$ 1024 & 167 & 0.36 \\
\bottomrule
\end{tabular}
\end{adjustbox}
}
\end{table}

\begin{figure}[!t]
  \centering
  \includegraphics[width=\linewidth]{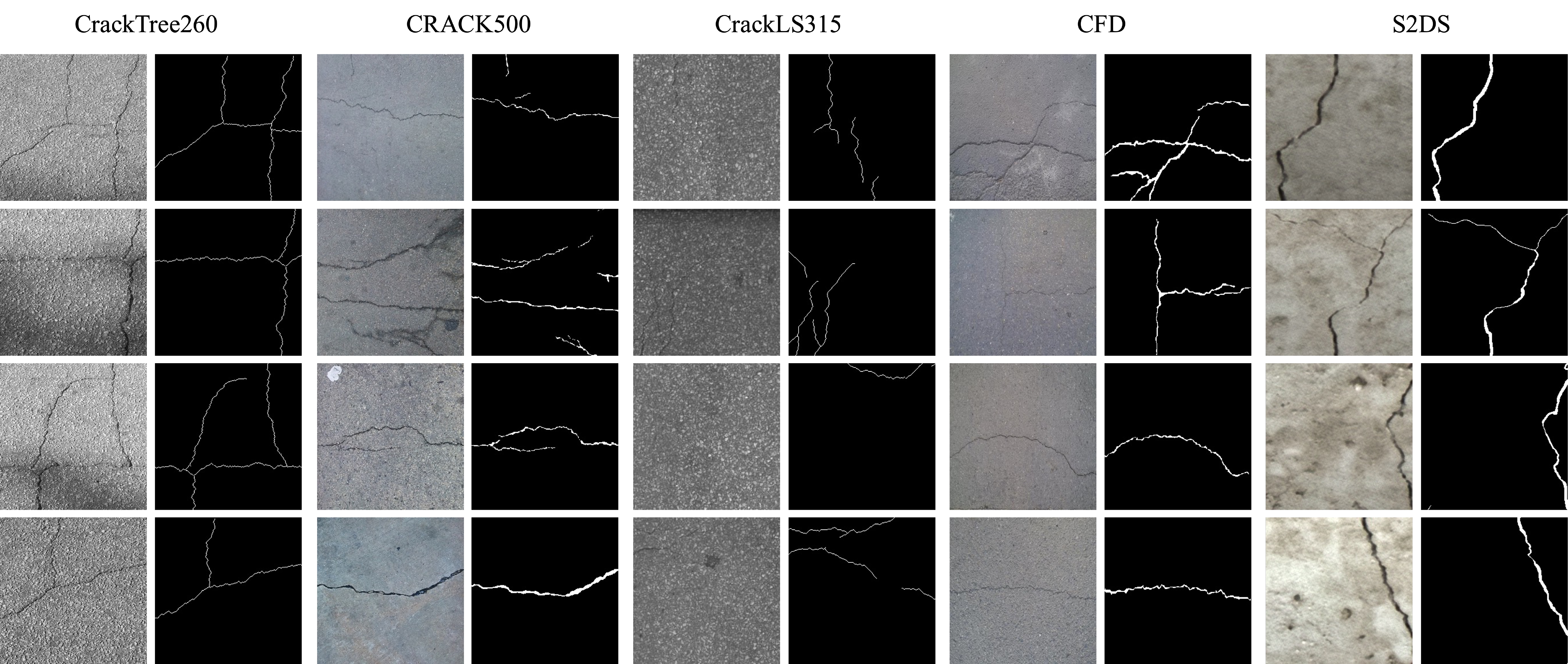}
  \caption{General display of the five datasets.}
  \label{fig:Exp_general}
\end{figure}

We provided some sample images and their corresponding masks in \cref{fig:Exp_general}. It is noticeable that the annotations of CrackTree260 and CrackLS315 have a width of 1 pixel, which is equivalent to the centerline. 
In practice, this 1-pixel convention creates severe class imbalance (typically $<\!1\%$ crack pixels), amplifying both FN (missed slender filaments) and FP (texture-driven false alarms) when training on small, domain-specific sets. The five datasets also differ in acquisition modality and background texture, which makes them an appropriate test bed for both in-domain accuracy and cross-domain robustness.

\subsection{Experimental setup}

\subsubsection{Implementation details}
We use stochastic (one-point) Flow Matching with a linear path \(x_t=(1-t)\,x_0 + t\,x_1\), where \(x_0\sim\mathcal{N}(0,I)\), \(x_1\) is a data sample, and \(t\sim\mathcal{U}(0,1)\). To enable classifier-free guidance (CFG), we randomly drop the semantic-mask conditioning with probability \(0.1\) (i.e., set \(y\) to zero); the mask generator uses the same schedule and drop strategy.

Across datasets we keep the same hyperparameters for both CrackSegFlow and the mask generator. At sampling, we use the guidance scale \(\omega=1.2\). Training uses Adam ~\cite{adam2014method} and an exponential moving average (EMA) of the weights with decay \(0.9999\). All experiments are implemented in PyTorch~2.8.0 and run on two NVIDIA A100 GPUs. Inputs are cropped/resized to $256{\times}256$. For each crack dataset, we curate 500 crack-centered patches and split them into $80/10/10$ for train/val/test; for CFD, we use its original 118 images (resized) and split them $80/20$ into train/test.

\subsubsection{Segmentation backbone}

As discussed in \cref{sec:related}, the dominant bottleneck in crack segmentation is cross-dataset generalization, not necessarily designing increasingly bespoke architectures. This is reflected in a very recent study \cite{goo2025hybrid}, which benchmarks several representative CNN and Transformer baselines using extensive augmentation on an aggregated crack dataset and reports only a modest gain from their hybrid CNN--Transformer model (mIoU $63.1\%$) over a classic U\!-\!Net (mIoU $60.8\%$).

Under a light hyperparameter-tuning protocol with standard data augmentation, we develop a compact baseline that pairs a U\!-\!Net decoder with a MiT-B4 (SegFormer-B4 \cite{xie2021segformer}) encoder, denoted as U\!-\!MiT. This model achieves mIoU $64.2\%$, exceeding their reported hybrid model while keeping architectural confounds minimal (\cref{tab:goo_iou}). The encoder uses the MiT-B4 hierarchical Transformer with overlapping patch embeddings to provide multi-scale context, while the U\!-\!Net decoder and skip connections preserve high-frequency detail critical for hairline cracks.

\FloatBarrier
\begin{table}[t]
\centering
\caption{mIoU (\%) on the dataset and test split of \cite{goo2025hybrid}. Baseline values (FCN through Hybrid-Segmentor) are taken from that work. We additionally report U\!-\!MiT, trained and evaluated on the same dataset and test split.}
\label{tab:goo_iou}
\setlength{\tabcolsep}{6pt}
\renewcommand{\arraystretch}{1.05}
{\footnotesize
\begin{tabular}{ccccccc}
\toprule
FCN & U\!-\!Net & DeepCrack2 & SegFormer & HrSegNet & Hybrid-Segmentor & U\!-\!MiT (Ours) \\
\midrule
59.9 & 60.8 & 59.0 & 56.8 & 59.5 & 63.1 & 64.2 \\
\bottomrule
\end{tabular}
}
\end{table}

In crack segmentation, foreground pixels are rare and thin, so the objective must be robust to class imbalance and sensitive to boundaries. We use a focal Tversky term \cite{abraham2019novel} for region supervision together with an edge-aware binary cross-entropy term, where the edge target is a Sobel-derived soft boundary map of the ground truth. Let $z\!\in\!\mathbb{R}^{1\times H\times W}$ be logits, $y\!\in\!\{0,1\}^{1\times H\times W}$ the ground truth, and $p=\sigma(z)$. With soft counts $\mathrm{TP}=\sum_i p_i y_i$, $\mathrm{FP}=\sum_i p_i(1-y_i)$, and $\mathrm{FN}=\sum_i (1-p_i)y_i$, the focal Tversky loss is
\begin{equation}
\label{eq:aftl}
T=\frac{\mathrm{TP}}{\mathrm{TP}+\alpha\,\mathrm{FP}+\beta\,\mathrm{FN}},
\qquad
\mathcal{L}_{\text{FT}}=(1-T)^{\gamma},
\end{equation}
with $(\alpha,\beta,\gamma)=(0.3,\,0.75,\,1.33)$, where $\beta>\alpha$ prioritizes recall of the minority crack class. The final objective combines region and boundary terms as
\begin{equation}
\label{eq:total}
\mathcal{L}=\lambda\,\mathcal{L}_{\text{FT}}+\eta\,\mathcal{L}_{\text{BCE}},
\qquad \lambda=0.8,\ \eta=0.2,
\end{equation}
with a short linear warm-up of the boundary term during early epochs for stability. In initial model selection, this objective consistently outperformed a binary cross-entropy plus Dice baseline \cite{du2023crack} across datasets and training settings.

All segmentation runs use U\!-\!MiT to evaluate the efficacy of CrackSegFlow and are trained for 100 epochs with AdamW, cosine decay with a short warm-up, separate encoder and decoder learning rates ($5\times10^{-5}$ and $5\times10^{-4}$), weight decay, and an effective global batch size of 64 via gradient accumulation. Overall, the goal of this paper is not to introduce another segmentation model, but to study synthesis and generalization under a stable, competitive backbone.

\subsubsection{Evaluation metrics.}
For the generative quality of synthesized masks and mask-conditioned rendered images produced by CrackSegFlow, we report Fr\'echet Inception Distance (FID) and Kernel Inception Distance (KID; reported as $\times 1000$), computed on Inception feature embeddings. Let $\Phi(\cdot)$ denote the Inception feature extractor, and let
$\mathcal{F}_{\text{real}}=\{\Phi(s)\,:\,s\in\mathcal{D}_{\text{real}}\}$ and
$\mathcal{F}_{\text{syn}}=\{\Phi(s)\,:\,s\in\mathcal{D}_{\text{syn}}\}$ be feature sets for real and synthetic samples (either masks or images). If $(\mu_{\text{real}},\Sigma_{\text{real}})$ and $(\mu_{\text{syn}},\Sigma_{\text{syn}})$ are the empirical mean and covariance of these features, FID is
\begin{equation}
\mathrm{FID}
=\left\|\mu_{\text{real}}-\mu_{\text{syn}}\right\|_{2}^{2}
+\mathrm{Tr}\!\left(\Sigma_{\text{real}}+\Sigma_{\text{syn}}
-2\left(\Sigma_{\text{real}}^{\frac{1}{2}}\,\Sigma_{\text{syn}}\,\Sigma_{\text{real}}^{\frac{1}{2}}\right)^{\frac{1}{2}}\right).
\label{eq:fid}
\end{equation}
KID measures the squared maximum mean discrepancy between $\mathcal{F}_{\text{real}}$ and $\mathcal{F}_{\text{syn}}$ under a polynomial kernel $k(\cdot,\cdot)$:
\begin{equation}
\mathrm{KID}
=\mathrm{MMD}^{2}(\mathcal{F}_{\text{real}},\mathcal{F}_{\text{syn}})
=\mathbb{E}\!\left[k(a,a')\right]+\mathbb{E}\!\left[k(b,b')\right]-2\mathbb{E}\!\left[k(a,b)\right],
\label{eq:kid}
\end{equation}
where $a,a'\sim\mathcal{F}_{\text{real}}$ and $b,b'\sim\mathcal{F}_{\text{syn}}$. Lower FID/KID indicates closer alignment between real and synthetic distributions.

For segmentation, we report mean Intersection-over-Union (mIoU) and F1 (Dice). Given a ground-truth mask $y\in\{0,1\}^{H\times W}$ and a thresholded prediction $\hat{y}\in\{0,1\}^{H\times W}$, define pixelwise $\mathrm{TP}$, $\mathrm{FP}$, and $\mathrm{FN}$. Then
\begin{equation}
\mathrm{IoU}=\frac{\mathrm{TP}}{\mathrm{TP}+\mathrm{FP}+\mathrm{FN}},
\qquad
\mathrm{F1}=\frac{2\,\mathrm{TP}}{2\,\mathrm{TP}+\mathrm{FP}+\mathrm{FN}}.
\label{eq:iou_f1}
\end{equation}
mIoU denotes the mean IoU over the evaluation set. IoU emphasizes overlap quality, while F1 balances precision and recall and is informative under extreme foreground sparsity.

\subsubsection{Synthesis protocol}

\begin{figure}[h!]
  \centering
  \includegraphics[width=0.8\linewidth]{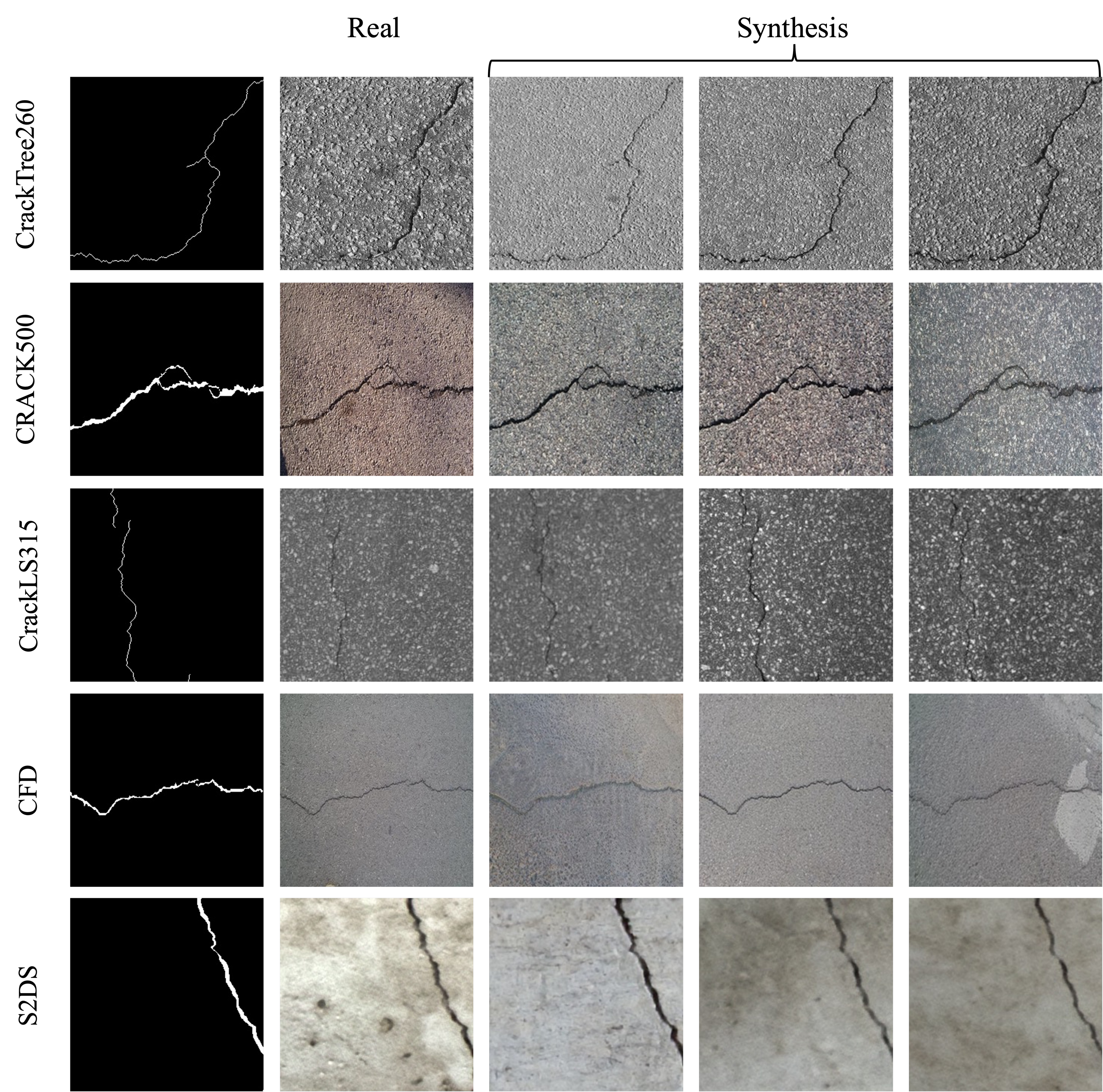}
  \caption{CrackSegFlow produces diverse images that strictly follow the provided semantic masks.}
  \label{fig:Exp_3synthesis}
\end{figure}

\begin{figure}[!t]
  \centering
  \includegraphics[width=0.8\linewidth]{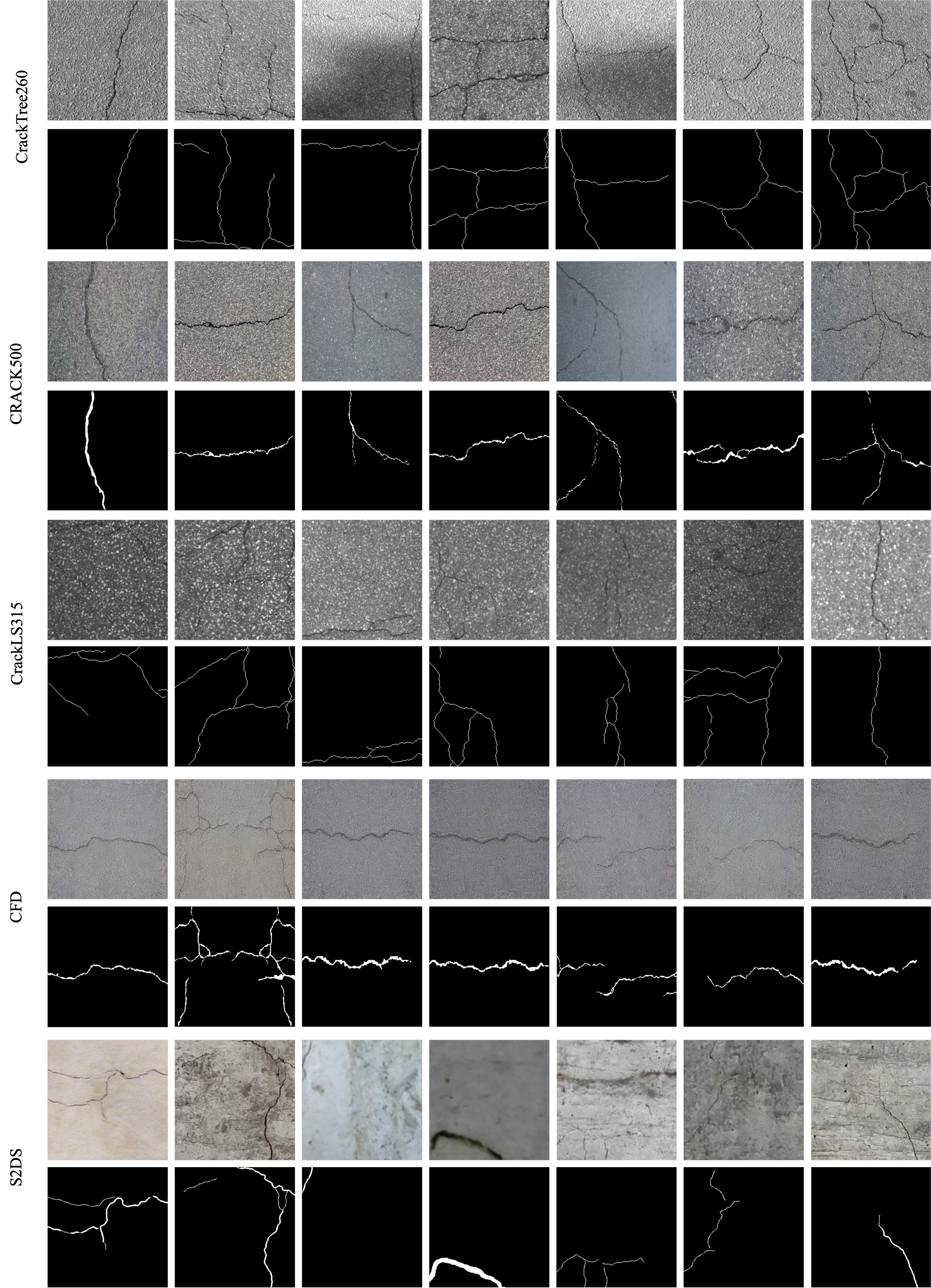}
  \caption{CrackSegFlow samples conditioned on synthetic masks from the mask generator.}
  \label{fig:Exp_allsynthesis}
\end{figure}

\paragraph{Synthesis quality}
As shown in ~\cref{fig:Exp_3synthesis}, CrackSegFlow generates images that are precisely aligned with the conditioning masks. To maximize mask diversity, we first sample semantic masks using the mask generator and then render images conditioned on these masks. As depicted in ~\cref{fig:Exp_allsynthesis}, the resulting images are novel in crack geometry and texture relative to the originals.~\cref{tab:fid_kid_three} summarizes generative quality for both modalities. Across datasets, Flow Matching achieves low FID/KID, with masks consistently scoring best (reflecting sharp topology), and images remaining competitive. These scores corroborate the visual fidelity in \Crefrange{fig:Exp_3synthesis}{fig:Exp_allsynthesis}
support the downstream gains, where higher-quality pairs correlate with improved mIoU/F1 after augmentation across all datasets.

\begin{table}[t]
\centering
\caption{Quality of Flow Matching synthesis measured by FID/KID (lower is better) for both mask and image modalities.}
\label{tab:fid_kid_three}
\setlength{\tabcolsep}{8pt}
{\footnotesize
\begin{tabular}{lccc}
\toprule
\textbf{Dataset} & \textbf{Modality} & \textbf{FID} $\downarrow$ & \textbf{KID $\times 1000$} $\downarrow$ \\
\midrule
\multirow{2}{*}{CrackTree260} & Image & 23.33 & 5.93 \\
                              & Mask  & 15.04 & 0.22 \\
\midrule
\multirow{2}{*}{CRACK500}     & Image & 28.94 & 10.38 \\
                              & Mask  & 23.66 & 4.44 \\
\midrule
\multirow{2}{*}{CrackLS315}   & Image & 21.76 & 7.19 \\
                              & Mask  & 13.61 & 0.11 \\
\midrule
\multirow{2}{*}{CFD}          & Image & 57.80 & 33.48 \\
                              & Mask  & 40.17 & 9.81 \\
\midrule
\multirow{2}{*}{S2DS}         & Image & 39.63 & 9.38 \\
                              & Mask  & 23.03 & 8.30 \\
\bottomrule
\end{tabular}
}
\end{table}

\paragraph{Synthesis policies for segmentation experiments}
For each dataset, let the real training set size be $x$.
For in-domain training, we sample $kx$ synthetic masks from the same dataset distribution and render $kx$ paired images conditioned on them. We use $k=16$ in the main setting.
For cross-domain target-guided transfer, for each source$\rightarrow$target pair, we inspect 10\% of the target training masks to estimate stable target mask statistics, including the crack-pixel-ratio distribution used to select sparsity bins. We then sample a synthetic mask set of size $4x_{\text{target}}$ from the mask generator conditioned on these target statistics, optionally apply lightweight width perturbations (controlled dilation/erosion), and render paired images using the source-trained CrackSegFlow renderer. The resulting pairs are used to train a syn-only model for that source$\rightarrow$target transfer.

\subsection{In-domain evaluation}

\begin{figure*}[t]
  \centering
  \includegraphics[width=\textwidth]{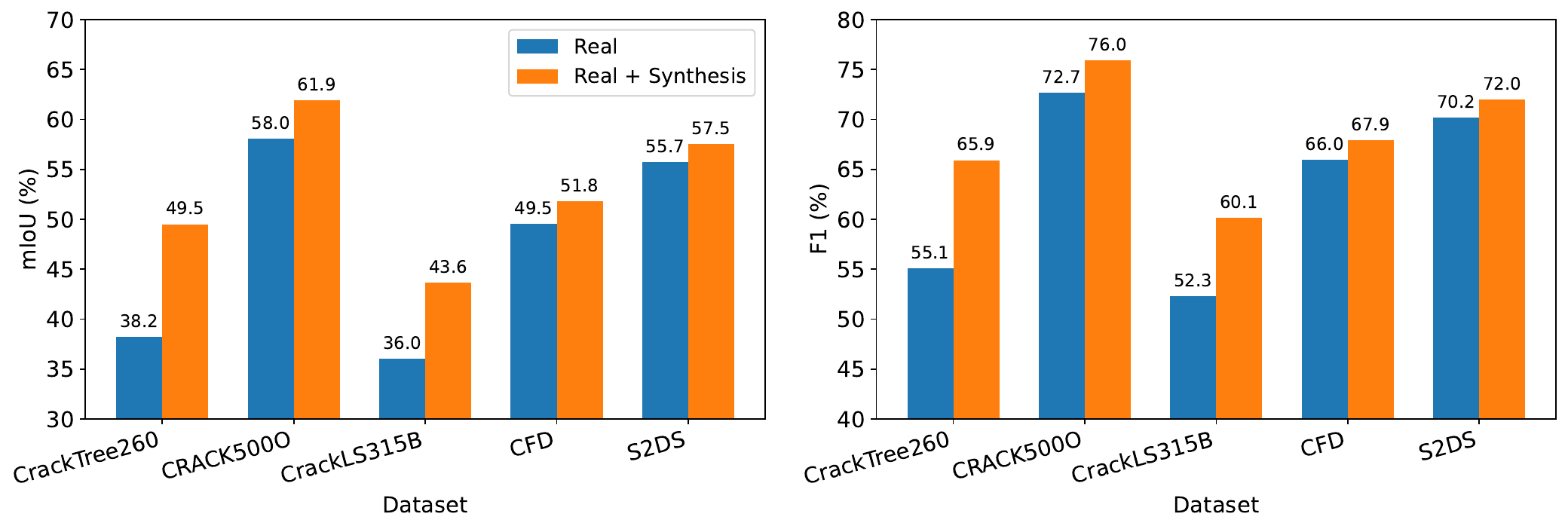}
  \caption{In-domain augmentation (+$16\times$ synthetic pairs) consistently improves mIoU and F1 across all datasets.}
  \label{fig:Exp_barchart}
\end{figure*}

In-domain results (see ~\cref{fig:Exp_barchart}) show that augmenting real training data with $16\times$ CrackSegFlow pairs (RS16) yields consistent gains across all five datasets under the same evaluation policy. Averaged over datasets, moving from real-only (R) to RS16 improves performance by \mbox{+5.37/+5.13} mIoU/F1, which corresponds to relative gains of \mbox{+13.0\%/+8.9\%}. The largest improvements occur on thin-crack benchmarks where continuity and width calibration are most challenging. CrackTree260 improves from \mbox{38.22/55.07} to \mbox{49.50/65.92} (absolute \mbox{+11.28/+10.85}, relative \mbox{+29.5\%/+19.7\%}), and CrackLS315B improves from \mbox{36.01/52.32} to \mbox{43.65/60.14} (absolute \mbox{+7.64/+7.82}, relative \mbox{+21.2\%/+14.9\%}). By comparison, gains on texture-diverse datasets are smaller but still reliable (for example, CRACK500 improves from \mbox{58.05/72.68} to \mbox{61.94/75.95}), indicating that mask-conditioned synthesis complements real imagery even when the baseline is strong.

\begin{figure*}[t]
  \centering
  \includegraphics[width=\textwidth]{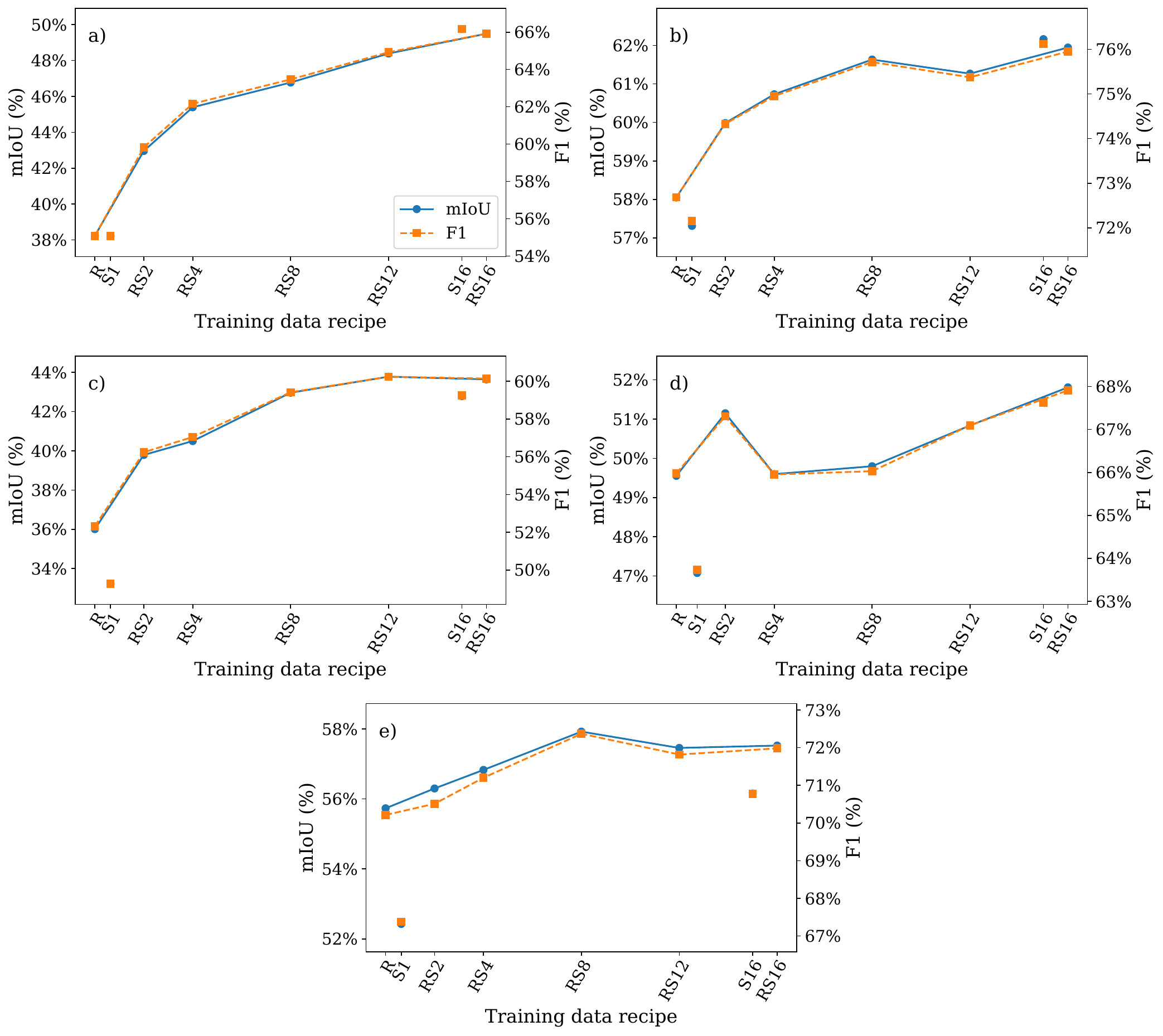}
  \caption{In-domain augmentation under the full synthesis-policy sweep across five datasets: (a) CrackTree260, (b) CRACK500, (c) CrackLS315, (d) CFD, (e) S2DS.}
  \label{fig:Exp_barchart2}
\end{figure*}

\begin{figure*}[t]
  \centering
  \includegraphics[width=0.95\textwidth]{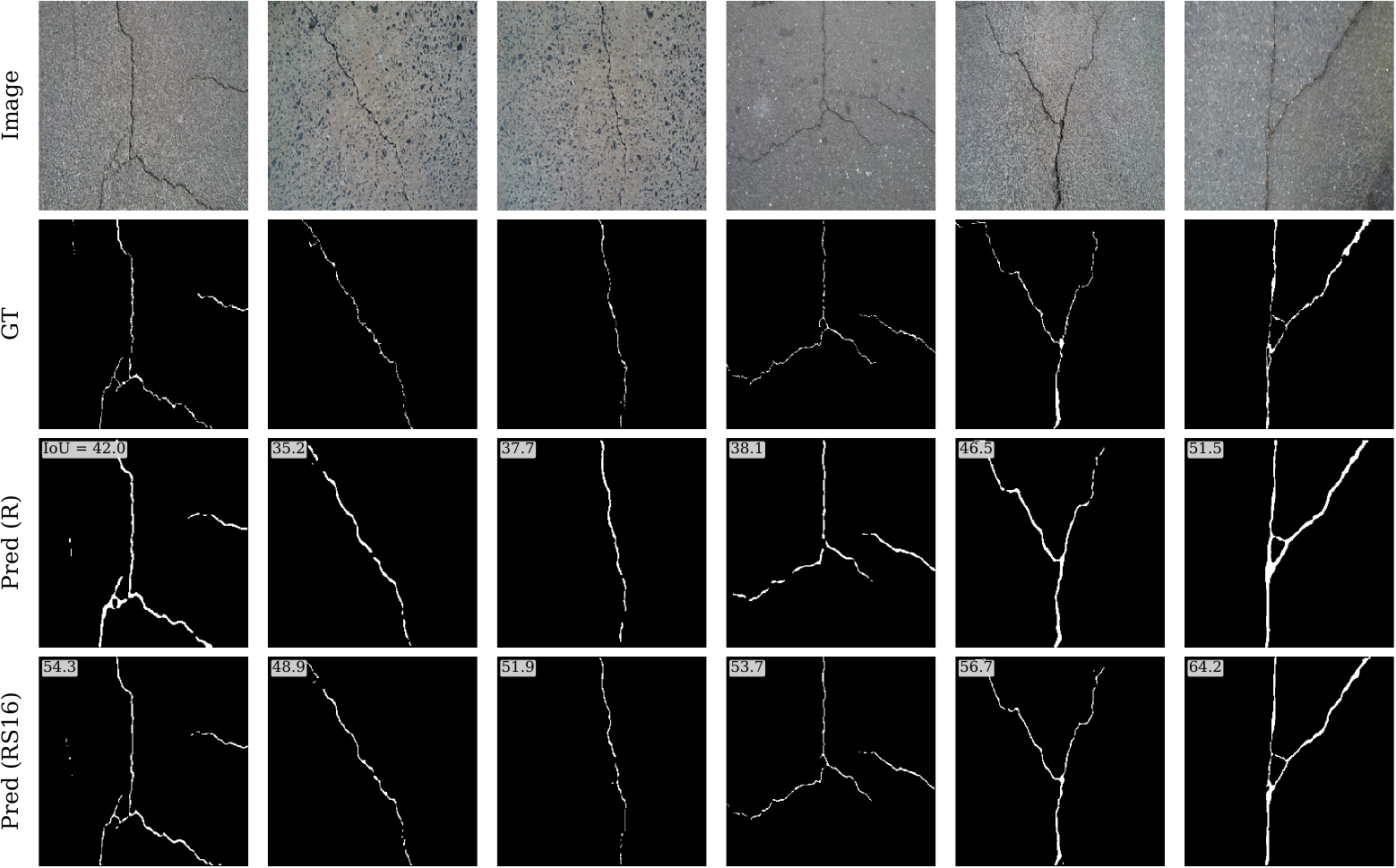}
  \caption{Qualitative comparison on CRACK500: real-only training vs.\ real+$8\times$ synthetic. Synthetic augmentation reduces over-thick predictions and aligns widths with ground truth.}
  \label{fig:Exp_in_CRACK500}
\end{figure*}

\begin{figure*}[t]
  \centering
  \includegraphics[width=0.95\textwidth]{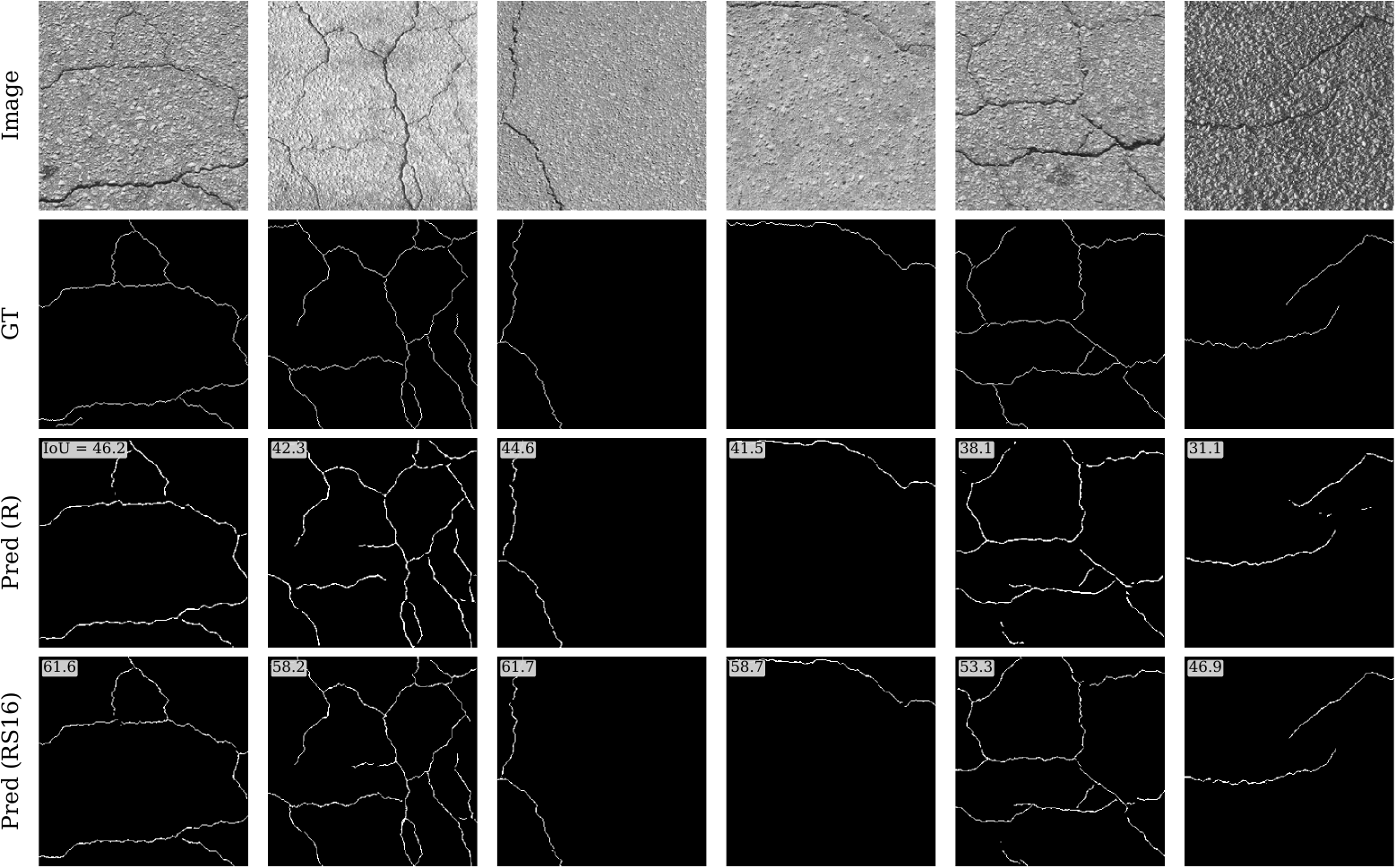}
  \caption{Qualitative comparison on CrackTree260: augmentation suppresses false positives while preserving 1-pixel-wide crack continuity.}
  \label{fig:Exp_in_CrackTree260}
\end{figure*}

The full policy sweep in ~\cref{fig:Exp_barchart2} reveals a consistent trend. Increasing the number of synthetic variants generally improves both synthetic-only (S$k$) and mixed (RS$k$) training, with diminishing returns beyond roughly 8--16 variants. Importantly, synthetic supervision can be competitive with real supervision when the generated pairs are topologically faithful. For CrackTree260, S1 essentially matches R (\mbox{38.22/55.08} versus \mbox{38.22/55.07}), and S16 reaches \mbox{49.76/66.17}, demonstrating that CrackSegFlow samples are sufficiently realistic to serve as primary training data in thin-structure regimes. Mixed training remains the most stable overall across datasets, with the best operating point depending on the domain (for example, RS12 on CrackLS315B at \mbox{43.78/60.24}, RS16 on CFD at \mbox{51.81/67.91}, and RS8 on S2DS at \mbox{57.92/72.37}). Qualitative comparisons in~\cref{fig:Exp_in_CRACK500,fig:Exp_in_CrackTree260}
corroborate these trends by showing reduced false positives on background texture, improved thin-crack connectivity, and better alignment between predicted crack width and the ground-truth centerlines. Additional results for the remaining datasets are provided in the Supplementary Material.

\subsection{Cross-domain evaluation}

Cross-domain crack segmentation is challenged by distribution shifts in surface texture, imaging conditions, and crack morphology, which reduce the transferability of models trained on a single source dataset. To address this with minimal target supervision, we use the target-guided synthesis protocol described above. Briefly, target crack statistics estimated from a small subset of target training masks guide the mask generator to produce target-like topology, and the source-trained CrackSegFlow renderer then produces aligned image--mask pairs under the source appearance prior. For each source$\rightarrow$target pair, we generate $4x_{\text{target}}$ synthetic pairs and train a syn-only transfer model. This couples target-consistent structure with realistic source photometrics, which is especially beneficial when cross-domain errors are dominated by morphological mismatch rather than model capacity.

~\cref{tab:cross_real5_miou_f1} reports cross-domain performance under real-only training, where each cell provides mIoU and F1 evaluated on the target test set using the source validation optimized threshold. The averages in the last column highlight that real-only transfer remains challenging across all sources. For example, the CFD source exhibits the lowest average performance, with 21.7 mIoU and 33.0 F1, indicating that a model trained on CFD struggles to generalize to other domains under the same thresholding policy. By contrast, CRACK500 attains the highest real-only average, 32.7 mIoU and 47.2 F1, suggesting that it provides comparatively stronger transferable priors. Even for this strongest source, however, the off-diagonal entries remain well below typical in-domain performance, underscoring the persistent impact of domain shift. The table also reveals pronounced pairwise asymmetries. A salient example is CFD to CrackTree260 and CFD to CrackLS315, both at 10.3 mIoU and 18.5 F1, which reflects severe under-segmentation and poor overlap when transferring from CFD into thin-crack domains with different background statistics. In contrast, CRACK500 to CFD achieves 47.7 mIoU and 64.0 F1, illustrating that certain sources transfer more effectively into CFD than others due to differences in texture diversity and crack appearance.

~\cref{tab:cross_aug5_miou_f1} summarizes the augmented setting using synthetic data, with the same evaluation protocol. The effect of target-guided synthesis is consistently positive across all sources and targets, and the magnitude of the gain is most pronounced in cross-domain settings with the largest baseline gap, supporting the central claim that the developed framework yields more important improvements in cross-domain generalization than in-domain refinement. The most substantial improvement appears for CFD as the source, where the average increases from 21.7 to 42.5 mIoU and from 33.0 to 58.1 F1, corresponding to absolute gains of 20.8 and 25.1 percentage points, respectively. These gains indicate that the synthesis guided by 10\% target mask statistics effectively corrects the dominant mismatch between CFD-trained crack priors and target crack topology. Additional large improvements are observed for CrackTree260 and CrackLS315 as sources, where averages increase from 30.1 to 44.3 mIoU and from 45.3 to 59.7 F1 for CrackTree260, and from 30.4 to 44.5 mIoU and from 45.8 to 60.5 F1 for CrackLS315. Notably, CRACK500 remains the strongest source on average after augmentation, reaching 41.0 mIoU and 56.8 F1, while also improving substantially relative to its real-only baseline. S2DS similarly improves from 26.2 to 34.4 mIoU and from 39.3 to 49.6 F1. Overall, the averages indicate that augmentation reduces cross-domain error in a broad and systematic manner rather than benefiting only a small subset of pairs.

\begin{figure*}[t]
  \centering
  \includegraphics[width=\textwidth]{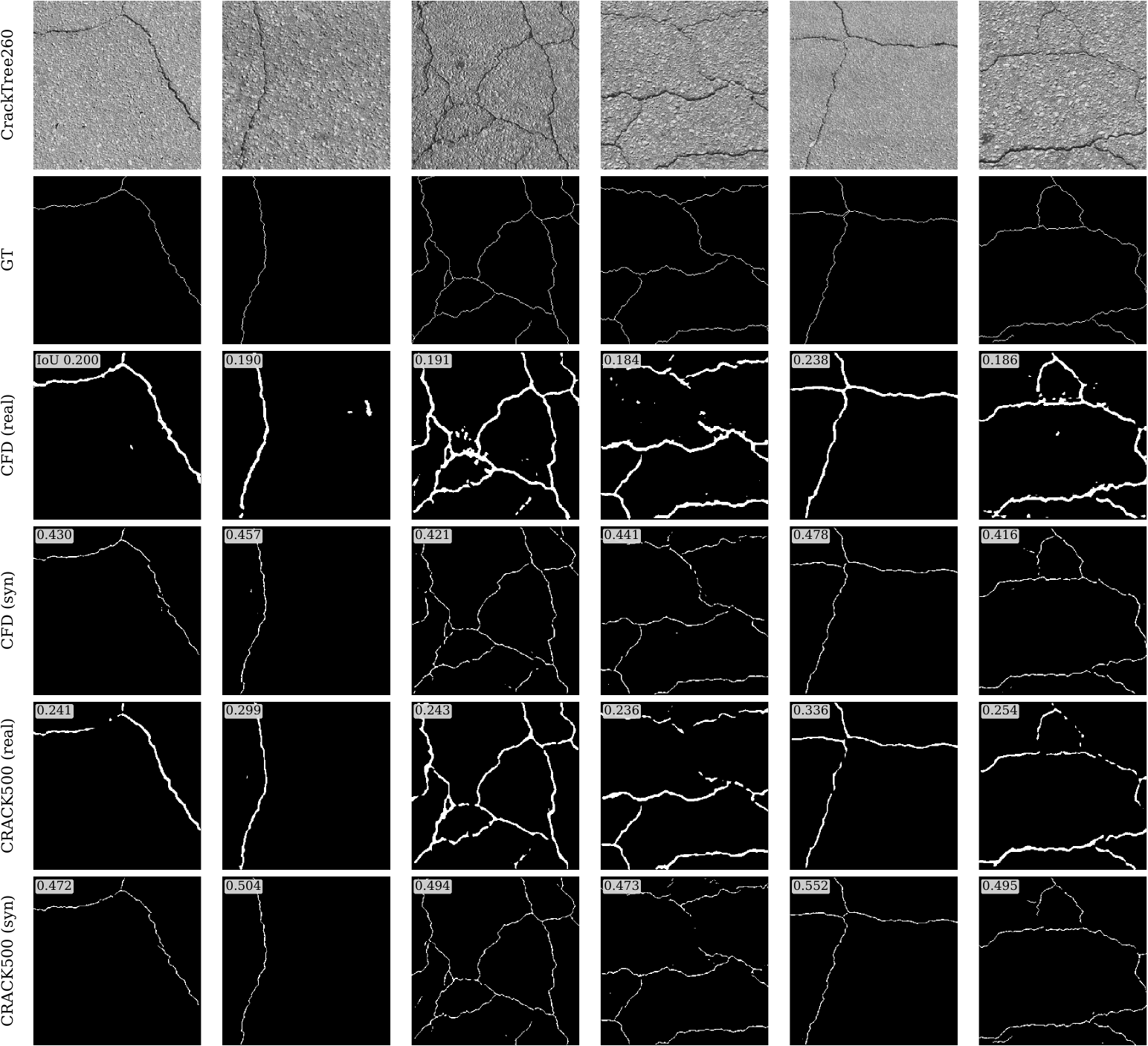}
  \caption{Cross-domain qualitative results on CrackTree260 using models trained on CFD and CRACK500. Rows show the target image, ground truth mask (cracks in white), and predictions from real-only and cross-domain syn-only training for each source. IoU is overlaid on the prediction rows for per-image comparison.}
  \label{fig:qual_cd_1}
\end{figure*}

The qualitative results in ~\cref{fig:qual_cd_1} provide instance-level evidence that complements the quantitative trends in ~\cref{tab:cross_real5_miou_f1,tab:cross_aug5_miou_f1}.~\cref{fig:qual_cd_1} evaluates CrackTree260 as the target using models trained on CFD and CRACK500. Consistent with ~\cref{tab:cross_real5_miou_f1}, the CFD real-only prediction exhibits fragmented crack responses and substantial missed detections, aligning with the low baseline of 10.3 mIoU and 18.5 F1 for CFD$\rightarrow$CrackTree260. After syn-only training with target-guided masks, the predicted structures become more continuous and topologically faithful, with fewer broken segments and improved connectivity at junctions, which is consistent with the corresponding improvement to 34.8 mIoU and 51.3 F1 in ~\cref{tab:cross_aug5_miou_f1}. The CRACK500 source shows a stronger baseline and further benefits from syn-only training, where the overlays indicate improved alignment and reduced spurious activations on background texture, supporting the increase from 22.8 and 37.0 to 37.7 and 54.3. In the Supplementary Material, additional qualitative results target S2DS using CFD and CrackLS315. The augmented models show clearer suppression of false positives on texture patterns and improved recovery of thin crack traces that are frequently under-segmented in the real-only setting. This qualitative behavior agrees with the strong numerical gains, particularly for CFD$\rightarrow$S2DS from 26.3 and 39.6 to 48.1 and 62.9. Further Supplementary examples target CRACK500 using CrackTree260 and CrackLS315 as sources. The syn-only rows exhibit smoother, more coherent crack networks with fewer isolated blobs and fewer missing branches, indicating simultaneous reductions in false positives and false negatives. The per-image IoU overlays corroborate these improvements and align with the large increases reported in ~\cref{tab:cross_aug5_miou_f1}, including CrackTree260$\rightarrow$CRACK500 from 32.8 and 48.4 to 55.5 and 70.5, and CrackLS315$\rightarrow$CRACK500 from 31.0 and 46.1 to 51.9 and 67.4. Taken together, the quantitative results and qualitative examples demonstrate that target-guided synthesis improves cross-domain segmentation by enforcing target-consistent crack topology while maintaining realistic appearance, yielding robust gains in both overlap quality and detection accuracy.

\newcommand{\mif}[2]{\mbox{#1\,/\,#2}}   
\newcommand{\dashcell}{\mbox{-}}        

\begin{table*}[t]
\centering
\caption{Cross-domain performance of real-only baselines. Cells report mIoU / F1 in percent on the target test set at the source validation optimized threshold. The last column reports the mean over targets, excluding the diagonal.}
\label{tab:cross_real5_miou_f1}
\setlength{\tabcolsep}{4.2pt}
\renewcommand{\arraystretch}{1.10}
{\footnotesize
\begin{tabular}{lcccccc}
\toprule
\textbf{source$\backslash$target} & \textbf{CrackTree260} & \textbf{CRACK500} & \textbf{CrackLS315} & \textbf{CFD} & \textbf{S2DS} & \textbf{Avg.} \\
\midrule
\textbf{CrackTree260} & \dashcell        & \mif{32.8}{48.4} & \mif{27.9}{42.8} & \mif{34.8}{51.3} & \mif{24.9}{38.7} & \mif{30.1}{45.3} \\
\textbf{CRACK500}     & \mif{22.8}{37.0} & \dashcell        & \mif{20.8}{33.9} & \mif{47.7}{64.0} & \mif{39.5}{53.7} & \mif{32.7}{47.2} \\
\textbf{CrackLS315}   & \mif{31.2}{47.4} & \mif{31.0}{46.1} & \dashcell        & \mif{34.5}{50.8} & \mif{24.9}{38.7} & \mif{30.4}{45.8} \\
\textbf{CFD}          & \mif{10.3}{18.5} & \mif{40.0}{55.5} & \mif{10.3}{18.5} & \dashcell        & \mif{26.3}{39.6} & \mif{21.7}{33.0} \\
\textbf{S2DS}         & \mif{16.7}{28.1} & \mif{37.9}{53.6} & \mif{14.0}{23.6} & \mif{36.2}{51.7} & \dashcell        & \mif{26.2}{39.3} \\
\bottomrule
\end{tabular}
}
\end{table*}

\begin{table*}[t]
\centering
\caption{Cross-domain performance of syn-only models trained on target-guided synthetic pairs. For each target, the synthetic set is four times the target size and is generated using source-trained CrackSegFlow with masks conditioned on 10 percent target statistics. Evaluation follows the same threshold protocol as~\cref{tab:cross_real5_miou_f1}.}
\label{tab:cross_aug5_miou_f1}
\setlength{\tabcolsep}{4.2pt}
\renewcommand{\arraystretch}{1.10}
{\footnotesize
\begin{tabular}{lcccccc}
\toprule
\textbf{source$\backslash$target} & \textbf{CrackTree260} & \textbf{CRACK500} & \textbf{CrackLS315} & \textbf{CFD} & \textbf{S2DS} & \textbf{Avg.} \\
\midrule
\textbf{CrackTree260} & \dashcell        & \mif{55.5}{70.5} & \mif{32.0}{47.8} & \mif{50.6}{67.0} & \mif{39.1}{53.3} & \mif{44.3}{59.7} \\
\textbf{CRACK500}     & \mif{37.7}{54.3} & \dashcell        & \mif{30.8}{46.1} & \mif{50.2}{66.4} & \mif{45.4}{60.4} & \mif{41.0}{56.8} \\
\textbf{CrackLS315}   & \mif{37.2}{54.0} & \mif{51.9}{67.4} & \dashcell        & \mif{45.3}{61.9} & \mif{43.5}{58.7} & \mif{44.5}{60.5} \\
\textbf{CFD}          & \mif{34.8}{51.3} & \mif{56.4}{71.5} & \mif{30.8}{46.5} & \dashcell        & \mif{48.1}{62.9} & \mif{42.5}{58.1} \\
\textbf{S2DS}         & \mif{26.3}{41.1} & \mif{47.0}{62.8} & \mif{21.7}{35.1} & \mif{42.7}{59.5} & \dashcell        & \mif{34.4}{49.6} \\
\bottomrule
\end{tabular}
}
\end{table*}

\subsection{Comparison to diffusion-based semantic synthesis}
\label{subsec:diffusion_comp}

CrackSegFlow is a pixel-domain semantic synthesis framework. Therefore, a fair comparison should address two diffusion-based alternatives that are commonly used for mask-guided image generation. The first is latent-diffusion conditioning, which is often adopted to accelerate sampling by operating in a learned latent space. The second is pixel-domain semantic diffusion, which matches our operating domain and enables an apple-to-apple evaluation.

\begin{figure}[t]
  \centering
  \includegraphics[width=\linewidth]{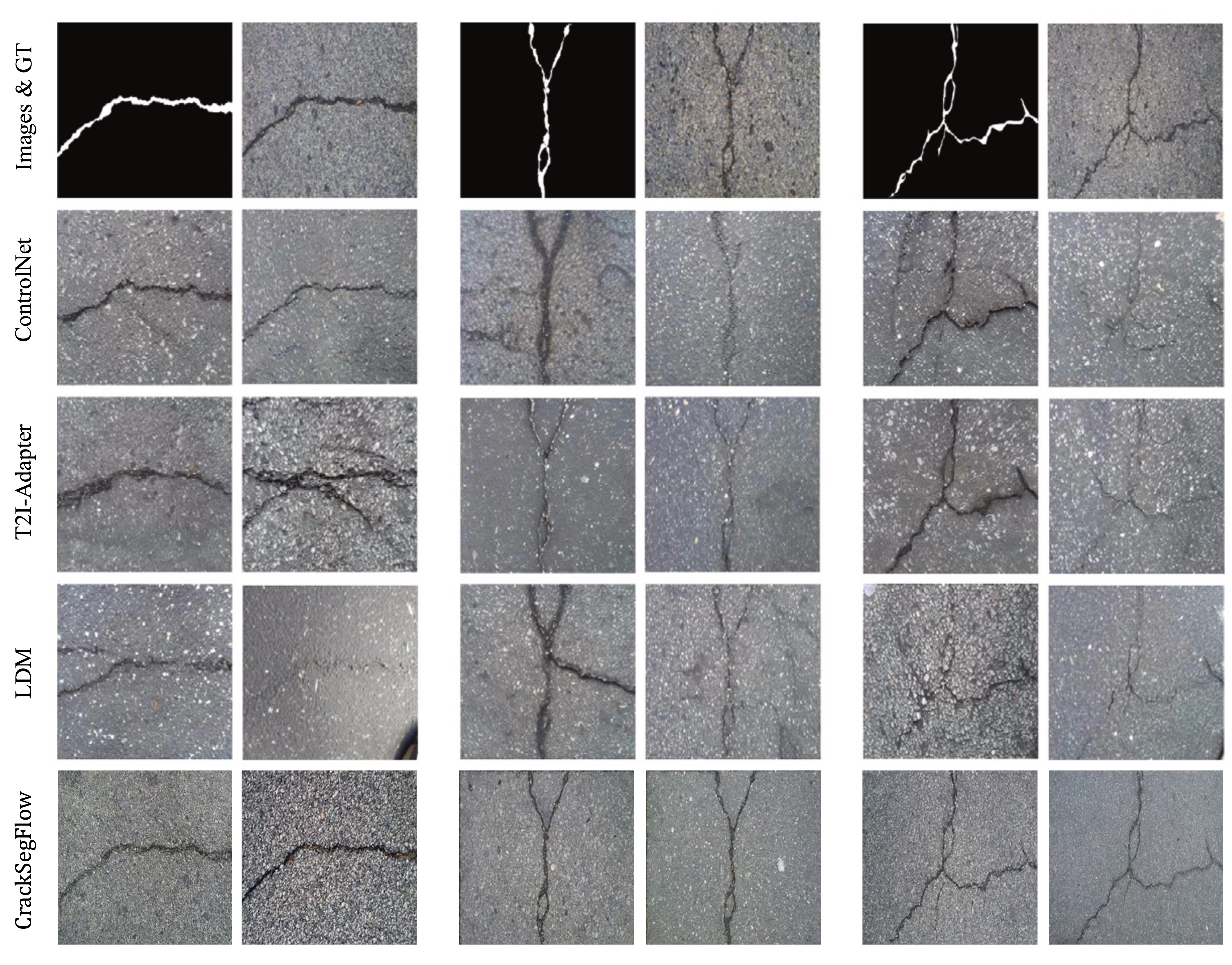}
 \caption{Comparison between latent-diffusion-based conditioning (LDM/ControlNet/T2I-Adapter) and CrackSegFlow for mask-guided crack synthesis. Latent conditioning may attenuate fine-scale geometry cues through normalization and feature mixing, whereas CrackSegFlow preserves thin-structure topology via persistent mask injection and deterministic transport. The latent-diffusion outputs are adapted from~\cite{lei2025faithful}.}

  \label{fig:ldm_vs_fm}
\end{figure}

Latent-diffusion pipelines, including LDM \cite{rombach2022ldm} and its conditioned variants such as ControlNet \cite{zhang2023adding} and T2I-Adapter \cite{mou2024t2i}, inject mask information through concatenation or auxiliary conditioning branches in the latent U-Net. While effective for general semantic layouts, this design can be sub-optimal for geometry-sensitive structures such as hairline cracks. In particular, repeated normalization and feature mixing can attenuate the conditioning signal at fine scales, leading to blurred filaments, weakened connectivity at junctions, and occasional mask--image drift~\cite{lei2025faithful}. \cref{fig:ldm_vs_fm} illustrates this limitation; latent-diffusion outputs often blur thin cracks or misalign them with the conditioning mask, whereas the FM formulation replaces stochastic noise injection with a continuous, condition-aware transport that preserves topology and alignment. CrackSegFlow avoids this failure mode by operating directly in the pixel space and preserving the conditioning signal through topology-preserving mask injection applied throughout the decoding part, coupled with deterministic transport that does not repeatedly corrupt intermediate states with stochastic noise. 

One may ask whether diffusion remains competitive when it is also implemented in the pixel domain. To answer this question, we train a semantic diffusion baseline following the semantic image synthesis formulation of Park et al. \cite{wang2022semantic} and compare it to CrackSegFlow under matched settings, including comparable training time and identical output resolution. In training, the diffusion baseline required approximately \(73\%\) more wall-clock time to reach the 120k to 150k step range in which CrackSegFlow already produced high-quality samples, with CRACK500 having FID of \(28.94\). Even after extending training to the 120k checkpoint, the diffusion model remained substantially worse, with FID of \(95.04\) under the recommended settings of the baseline.

\begin{figure}[t]
  \centering
  \includegraphics[width=0.9\linewidth]{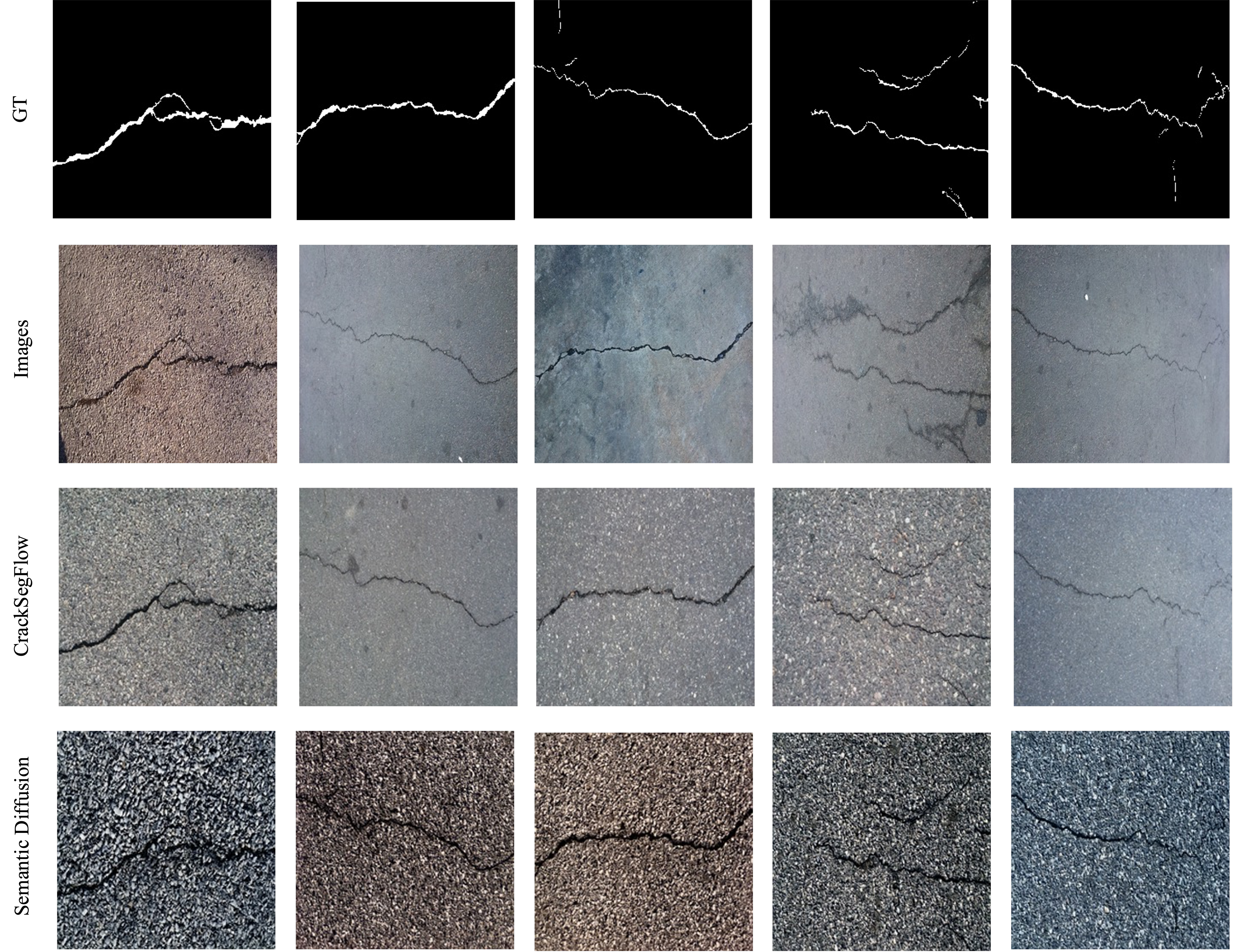}
  \caption{Apple-to-apple comparison in the pixel domain between the best semantic diffusion baseline and a matched checkpoint of CrackSegFlow under the same training setup and comparable compute. Diffusion outputs exhibit background texture artifacts and substantially lower sample fidelity.}
  \label{fig:Exp_diffusion_vs_flow_matching}
\end{figure}

At inference, despite both methods operating in the pixel domain, CrackSegFlow is approximately \(7\times\) faster at sampling than the diffusion baseline due to deterministic ODE sampling with a small number of integration steps rather than iterative denoising. ~\cref{fig:Exp_diffusion_vs_flow_matching} shows qualitative results from the best diffusion checkpoint and a matched CrackSegFlow checkpoint. The diffusion outputs exhibit noticeable background texture artifacts, whereas CrackSegFlow produces cleaner textures while adhering more faithfully to the intended crack topology.

\subsection{CSF-50K benchmark}
\label{subsec:csf50k}

We release CSF-50K, a fully synthetic benchmark of 50,000 paired crack images and pixel-accurate masks for training and evaluating generalizable crack segmentors. Compared with common public crack datasets (typically a few hundred images), CSF-50K is one to two orders of magnitude larger and is constructed to emphasize thin-structure topology, strict mask--image alignment, and appearance diversity. Importantly, as shown in ~\cref{fig:Exp_barchart}, training with our synthetic pairs is competitive with (and complementary to) real images, supporting the use of these data as effective supervision rather than only qualitative augmentation. The dataset is generated by combining three synthesis strategies developed in this work:

 \begin{figure}[t]
  \centering
  \includegraphics[width=0.85\linewidth]{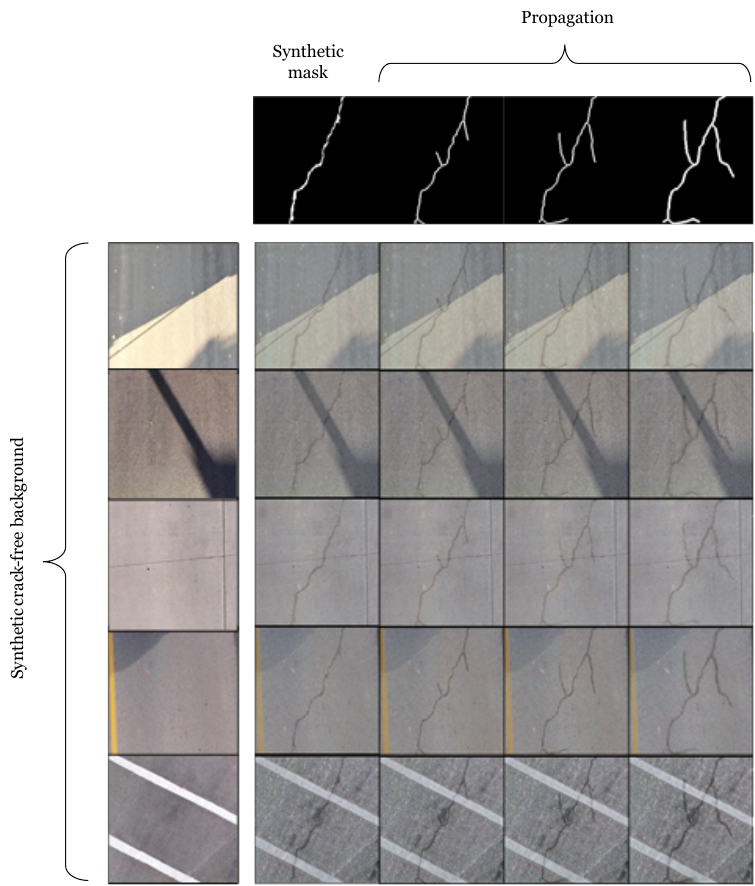}
  \caption{Background-guided crack injection. Given a crack-free background and a target crack mask, our rectified-flow renderer injects a realistic crack instance that strictly follows the mask while preserving the original background context.}
  \label{fig:bg_mask}
\end{figure}

\begin{itemize}
  \item Strategy A: mask-generator-driven paired synthesis (about 27k pairs). 
  For each of the five datasets (CRACK500, CrackTree260, CrackLS315, CFD, and S2DS), we first sample synthetic masks from the class-conditional FM mask generator by selecting sparsity bins that span ultra-sparse to denser crack regimes. We then render photorealistic images conditioned on these masks using CrackSegFlow, producing topology-consistent pairs as illustrated in ~\cref{fig:Exp_allsynthesis}. This strategy scales each dataset by a large multiplicative factor while keeping crack continuity and junction structure aligned to the conditioning masks.

  \item Strategy B: propagated-mask synthesis for intra-regime topology broadening (about 5k pairs).
  Starting from base masks, we generate structure-preserving propagated variants via controlled morphological perturbations (see ~\cref{fig:mask_propagation}) to broaden geometry within the same sparsity regime. Images are then rendered from these propagated masks, which increases diversity in width conventions and local branching patterns.

    \item Strategy C: background-guided crack injection (about 18k pairs).
  This strategy is designed to increase appearance diversity (lighting, texture, and imaging artifacts) and to reduce texture-driven false positives caused by crack-like patterns such as shadows, joints, and pavement markings. Since these non-crack structures often resemble thin cracks and frequently trigger spurious detections, we use intact pavement backgrounds that contain such patterns while keeping the background regions labeled as negative. We collected about 1,000 crack-free drone images in Urbana--Champaign and Rantoul, Illinois, and trained an unconditional FM background generator to sample additional crack-free backgrounds with similar variability, including changes in illumination, shadows, stains, and markings. We then sample masks using the class-conditional generator and propagation, and render crack images conditioned jointly on the sampled background and the target mask (see ~\cref{fig:bg_mask}). By repeatedly exposing the model to shadow/marking patterns as background (non-crack) while injecting cracks only where the mask indicates, this strategy directly teaches the segmentor to distinguish cracks from common confounders and decreases false positives.
\end{itemize}

CSF-50K is randomly split into 40k/5k/5k for training/validation/testing, and is available at
\href{https://github.com/BabakAsadi94/CrackSegFlow}{Dataset link}.

\section{Conclusion}
Crack segmentation is a foundational step in automated visual inspection pipelines (e.g., UAV/robot/vehicle imaging) used for condition assessment of pavements and other civil infrastructure, and it directly supports downstream maintenance prioritization and asset-management decisions. By reducing pixel-level labeling burden and improving cross-domain robustness through controllable, mask-aligned synthesis, the proposed framework strengthens the reliability and scalability of automated inspection systems deployed across sites, sensors, and surface textures.
This paper introduced CrackSegFlow, a controllable Flow Matching framework that renders realistic pavement images conditioned on an input crack mask while maintaining tight mask--image alignment. CrackSegFlow consists of two rendering modules---a topology-preserving mask injection module and a boundary-gated rendering module---that jointly enforce structural fidelity (connectivity and thin-structure continuity) and suppress mask-boundary artifacts, enabling the generated image to accurately match the provided mask under diverse pavement appearances. Because CrackSegFlow is explicitly mask-conditioned, the conditioning mask can be either real or synthetic, providing a unified mechanism for controllable data synthesis.

For mask construction, we adopted a class-conditional mask generation strategy, where the class is defined by crack pixel ratio using the same policy-level grouping as the experiments. This design yields direct control over crack coverage and enables stable scaling to multiple synthetic variants per image. In addition, we introduced a propagation strategy to diversify crack morphology by extending and expanding existing crack structures in a topology-consistent manner. To further diversify background illumination and texture---and to reduce false positives caused by background patterns---we incorporated a mask injection into crack-free backgrounds mechanism, where arbitrary masks are injected onto crack-free surfaces and rendered via rectified-flow to produce realistic crack appearance under varied photometric conditions.

To evaluate CrackSegFlow under a strong and consistent segmentation backbone, we developed a U-Net--style segmentor with a Transformer encoder (MiT-B4) together with a hybrid BCE + focal Tversky loss. While preliminary comparisons indicated that this backbone outperforms representative hybrid CNN--Transformer baselines, the main experiments did not aim to introduce a new segmentation architecture. Instead, the focus remained on how controllable Flow Matching synthesis improves (i) in-domain performance, (ii) cross-domain transfer, and (iii) practical scalability relative to diffusion-based augmentation. The main conclusions of this study are as follows:
\begin{itemize}
  \item Under in-domain training, multi-variant synthesis yields consistent improvements across all five datasets. For example, CrackTree260 increases from \mbox{38.22/55.07} to \mbox{49.50/65.92}, and CrackLS315B increases from \mbox{36.01/52.32} to \mbox{43.65/60.14} (mIoU/F1). Averaged over all datasets, the mean performance improves from \mbox{47.51/63.25} to \mbox{52.89/68.38}, corresponding to average absolute gains of \mbox{+5.37/+5.13} points (mIoU/F1), corresponding to relative gains of \mbox{+13.0\%/+8.9\%}

  \item Cross-domain transfer benefits substantially from target-guided synthesis with minimal target supervision. Using only 10\% target masks to estimate stable crack-geometry statistics and training syn-only models on target-guided synthetic sets (sized at 4$\times$ the target) increases the overall cross-domain average from \mbox{28.22/42.12} to \mbox{41.34/56.94}, corresponding to absolute gains of \mbox{+13.12/+14.82} points and relative gains of \mbox{+46.5\%/+35.2\%} (mIoU/F1). The gains are largest when real-only transfer is weakest; for example, using CFD as the source increases the source-wise average from \mbox{21.7/33.0} to \mbox{42.5/58.1}, and CrackTree260$\rightarrow$CRACK500 improves from \mbox{32.8/48.4} to \mbox{55.5/70.5}.

  \item Relative to both pixel- and latent-diffusion augmentation, CrackSegFlow provides a more practical operating point: deterministic sampling enables substantially faster generation while maintaining high perceptual fidelity (low FID/KID) and preserving tight mask--image alignment, which is critical for thin-structure supervision where small boundary errors directly translate into segmentation noise.

  \item Finally, the proposed pipeline scales to large, diverse supervision, and we release a 50k mask--image pair benchmark (CSF-50K) to support reproducible evaluation of controllable crack synthesis and cross-domain crack segmentation.
\end{itemize}

In future work, we plan to extend the framework established in this paper to additional infrastructure distresses (e.g., potholes and surface spalling) and to study multi-distress controllable synthesis for improved field robustness. Moreover, while the main focus in this paper is controllable synthesis rather than segmentation-model development (and our U\!-\!MiT backbone provides a solid evaluation baseline), we note that in ongoing work we are developing an FM-based crack segmentation model that shows strong promise on thin-structure benchmarks.

\section*{Acknowledgment} The first author gratefully acknowledges support from the University of Illinois Urbana-Champaign Graduate College through the Dissertation Completion Fellowship for the 2025--2026 academic year, awarded as the Department of Civil and Environmental Engineering recipient. The first author also thanks Professor Svetlana Lazebnik for helpful suggestions and feedback that improved the presentation of this work. This work used Delta GPU at the National Center for Supercomputing Applications (NCSA) through allocation CIS240456 from the Advanced Cyberinfrastructure Coordination Ecosystem: Services \& Support (ACCESS) program, which is supported by U.S. National Science Foundation grants \#2138259, \#2138286, \#2138307, \#2137603, and \#2138296.

\bibliographystyle{elsarticle-num}
\bibliography{ref}
\end{document}